\documentclass[journal=jacsat,manuscript=article]{achemso}

\usepackage{chemformula} 
\usepackage[T1]{fontenc} 
\usepackage{adjustbox}
\usepackage{multirow}
\usepackage{booktabs} 
\usepackage{amssymb} 
\usepackage[table]{xcolor}
\usepackage{rotating}
\usepackage{float}
\usepackage{hyperref}
\usepackage{enumitem}
\newcommand{\mname}{QCBench}
\usepackage[normalem]{ulem}
\usepackage{xcolor}

\author{Jiaqing Xie}

\affiliation[PJLAB]{Shanghai Artificial Intelligence Laboratory, 701 Yunjin Road, Xuhui, Shanghai, 200232, China}
\altaffiliation{These authors contributed equally to this work.}
\author{Weida Wang}
\affiliation[FDU]{Fudan University, 220 Handan Rd, Yangpu, Shanghai, 200433, China}
\altaffiliation{These authors contributed equally to this work.}
\alsoaffiliation[PJLAB]{Shanghai Artificial Intelligence Laboratory, 701 Yunjin Road, Xuhui, Shanghai, 200232, China}
\author{Ben Gao}
\affiliation[WHU]{Wuhan University, G9P7+CP8, Wuhan, Hubei, 430072, China}
\altaffiliation{These authors contributed equally to this work.}
\alsoaffiliation[PJLAB]{Shanghai Artificial Intelligence Laboratory, 701 Yunjin Road, Xuhui, Shanghai, 200232, China}

\author{Zhuo Yang}
\affiliation[XDU]{Xidian University, 266 Xinglong Section of Xifeng Road, Xi’an, Shaanxi, 710126, China}
\affiliation[PJLAB]{Shanghai Artificial Intelligence Laboratory, 701 Yunjin Road, Xuhui, Shanghai, 200232, China}
\author{Haiyuan Wan}
\affiliation[THU]{Tsinghua University, Haidian  District, Beijing, 100084, China}
\alsoaffiliation[PJLAB]{Shanghai Artificial Intelligence Laboratory, 701 Yunjin Road, Xuhui, Shanghai, 200232, China}
\author{Shufei Zhang}
\affiliation[PJLAB]{Shanghai Artificial Intelligence Laboratory, 701 Yunjin Road, Xuhui, Shanghai, 200232, China}
\author{Tianfan Fu}
\affiliation[NJU]{Nanjing University, 163 Xianlin Road, Qixia District, Nanjing, Jiangsu, 210023, China}
\alsoaffiliation[PJLAB]{Shanghai Artificial Intelligence Laboratory, 701 Yunjin Road, Xuhui, Shanghai, 200232, China}
\author{Yuqiang Li}
\affiliation[PJLAB]{Shanghai Artificial Intelligence Laboratory, 701 Yunjin Road, Xuhui, Shanghai, 200232, China}
\email{liyuqiang@pjlab.org.cn}

\title[An \textsf{achemso} demo]
  {QCBench: Evaluating Large Language Models on Domain-Specific Quantitative
Chemistry}

\abbreviations{IR,NMR,UV}
\keywords{American Chemical Society, \LaTeX}

\begin{document}







\begin{abstract}
Quantitative chemistry is central to modern chemical research, yet the ability of large language models (LLMs) to perform its rigorous, step-by-step calculations remains underexplored.
To fill this blank, we propose \mname, a \textbf{Q}uantitative \textbf{C}hemistry oriented benchmark comprising 350 computational chemistry problems across 7 chemistry subfields, which contains analytical chemistry, bio/organic chemistry, general chemistry, inorganic chemistry, physical chemistry, polymer chemistry and quantum chemistry. To systematically evaluate the mathematical reasoning abilities of large language models (LLMs), they are categorized into three tiers: easy, medium, and difficult. Each problem, rooted in realistic chemical scenarios, is structured to prevent heuristic shortcuts and demand explicit numerical reasoning. \mname\ enables fine-grained diagnosis of computational weaknesses, reveals model-specific limitations across difficulty levels, and lays the groundwork for future improvements such as domain-adaptive fine-tuning or multi-modal integration. Evaluations on 24 LLMs demonstrate a consistent performance degradation with increasing task complexity, highlighting the current gap between language fluency and scientific computation accuracy. Code for QCBench is available at \href{https://github.com/jiaqingxie/QCBench}{https://github.com/jiaqingxie/QCBench}. 
\end{abstract}

\section{Introduction}
\begin{figure}[!htb]
    \centering
    \includegraphics[width=0.7\linewidth]{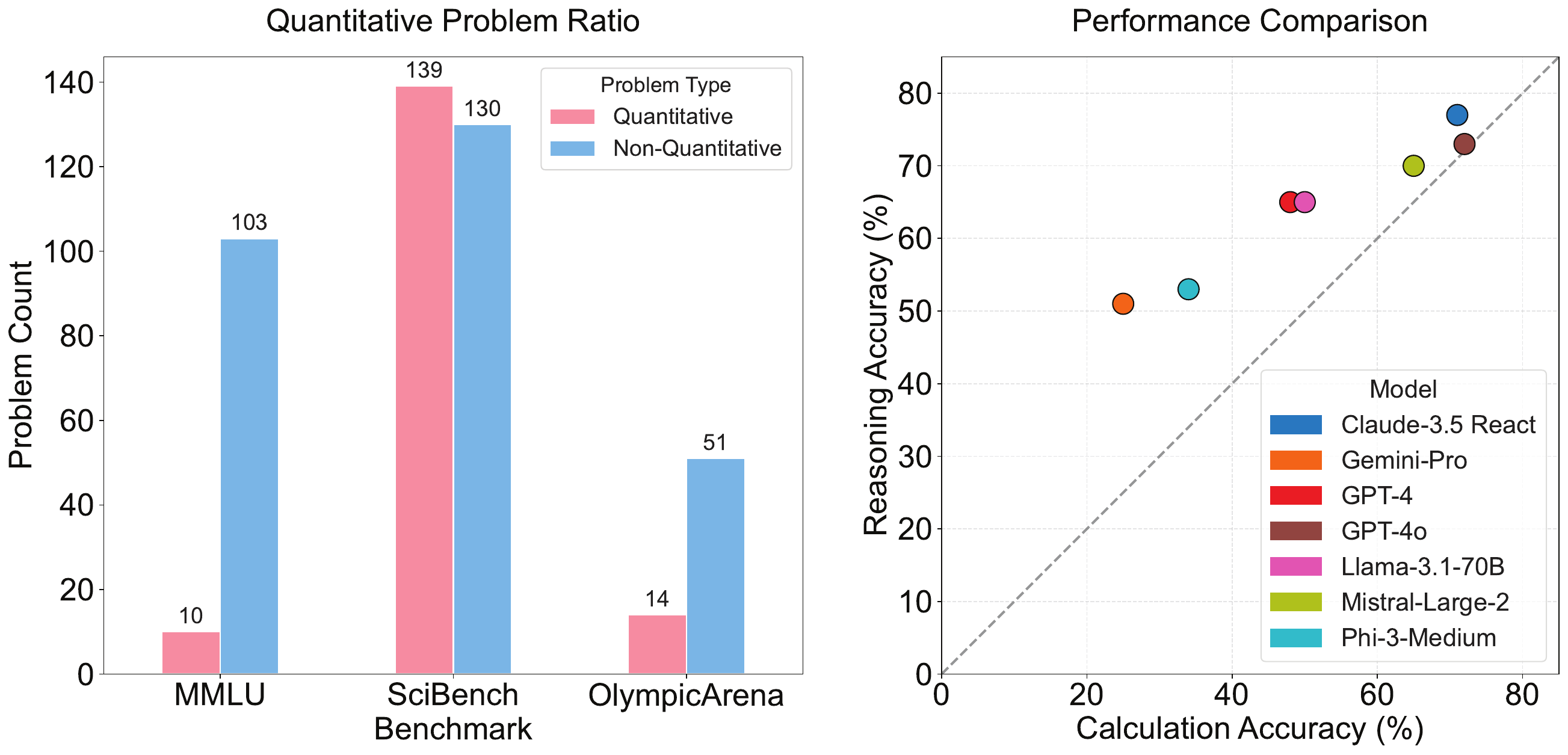}
    \vspace{-0.2cm}
    \caption{Quantitative questions ratio is relatively small in some chemistry benchmarks. Also, performance is relatively high for reasoning tasks as observed by ChemBench~\cite{mirza2024large}, which motivates us to curate a pure computing benchmark. Original figure created by the authors; not previously published.}
    \label{fig:quant_ratio}
\end{figure}

Numerous benchmark datasets have been proposed to evaluate the problem-solving capabilities of AI models across scientific disciplines, including mathematics \cite{gao2024omni, fan2024hardmath, glazer2024frontiermath}, physics \cite{chung2025theoretical, qiu2025phybench, he2024olympiadbench}, and chemistry \cite{wang2023scibench, guo2023can, mirza2024large}. In the domains of mathematics and physics, benchmark problems predominantly focus on theorem proving and numerical computation. However, analytical or comprehension-based problems, which require deeper semantic understanding, are underrepresented in mainstream math and physics datasets \cite{yue2024harp, yan2024survey, chung2025theoretical, zhang2025abench, qiu2025phybench}. In contrast, chemistry benchmarks often involve complex reasoning tasks that rely heavily on structural interpretation and image-based analysis. For instance, many problems require interpreting H-NMR spectra \cite{wang2025nmrextractor}, evaluating molecular properties from structural diagrams \cite{guo2024can}, or performing reaction synthesis prediction \cite{zhao2024chemdfm, mirza2024large, guo2023can}. Despite this diversity, current chemistry benchmarks remain limited in their breadth and depth. As shown in Figure~\ref{fig:quant_ratio}, only a small fraction of tasks in general-purpose benchmarks such as MMLU \cite{hendrycks2020measuring} and OlympicArena \cite{huang2024olympicarena} are computational in nature, constituting merely 10 and 14 problems, respectively, while over 80\% are categorized as non-quantitative.
\begin{figure*}[!htb]
    \centering
    \includegraphics[width=1\linewidth]{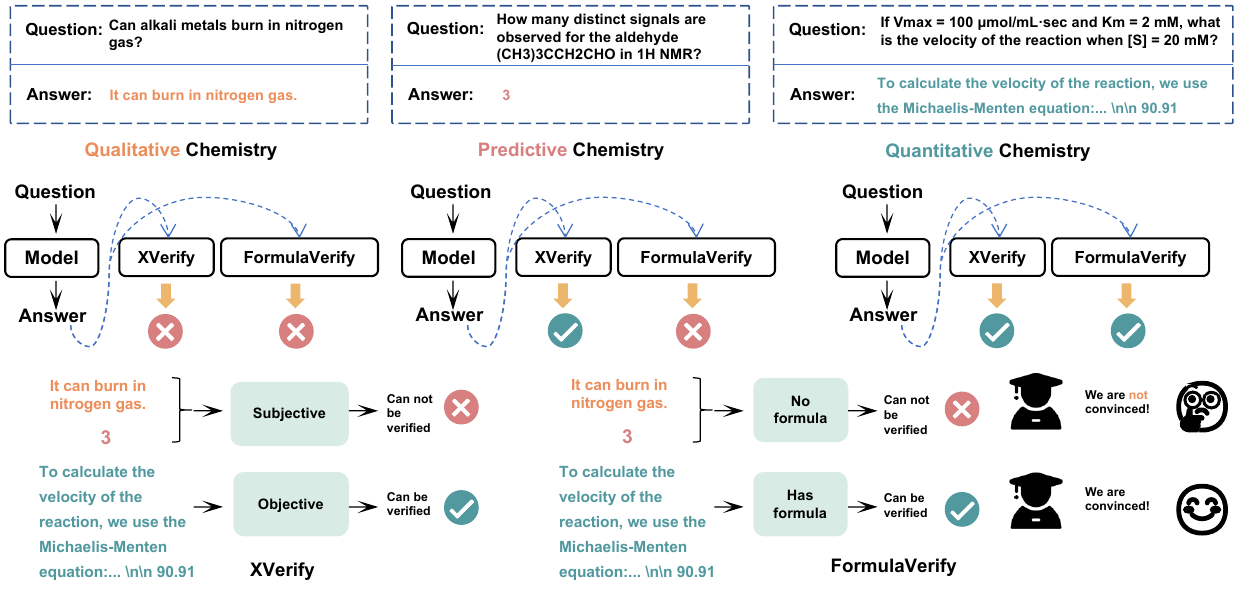}
    \caption{\textit{Qualitative chemistry} versus \textit{predictive chemistry} versus \textit{quantitative chemistry}. We define \textit{quantitative} chemistry problems as those requiring objective, numerical answers that can be automatically evaluated by tools such as xVerify \cite{chen2025xverify}. In contrast, qualitative problems involve textual explanations or conceptual reasoning without definitive ground-truth answers. \textit{Predictive} chemistry includes simple counting tasks, such as tallying hydrogen atoms, or interpreting reaction equations, as they lack formal computational steps and do not require formula-based reasoning.}
    \label{fig:define_quantative_chemistry}
\end{figure*}
A notable limitation of existing chemistry benchmarks lies in the incomplete and often ambiguous categorization of problem domains. For instance, SciBench~\cite{wang2023scibench} includes tasks primarily drawn from \textit{Quantum Chemistry} and \textit{Physical Chemistry}, while ChemBench~\cite{mirza2024large} encompasses seven chemistry domains. However, upon closer inspection, the \textit{Analytical Chemistry} problems in ChemBench predominantly revolve around tasks such as determining the number of hydrogen atoms from H-NMR spectra~\cite{wang2025nmrextractor}. We claim that such tasks should be categorized as \textbf{predictive} \textit{rather than} \textbf{quantitative} (Figure~\ref{fig:define_quantative_chemistry}). Truly quantitative problems in scientific domains involve explicit computational reasoning, typically requiring the application of formulas, step-by-step calculations (\textit{e.g.}, equations, numerical approximations), and yielding numerical outputs, often as floating-point values \cite{pang2006introduction}. Furthermore, tasks such as chemical equation balancing, although appearing procedural, do not constitute genuine quantitative reasoning. They primarily test conceptual understanding of chemical principles, such as the conservation of mass, rather than mathematical computation. Similarly, predicting the number of structural isomers or determining electron counts in molecules involves rule-based inference or pattern recognition, and should not be conflated with quantitative problem solving. This misclassification leads to a second drawback: many so-called computational tasks in existing benchmarks do not reflect the depth of quantitative reasoning expected in math or physics domains. As shown in the right panel of Figure~\ref{fig:quant_ratio}, this discrepancy is further reflected in model performance. Current state-of-the-art models demonstrate stronger reasoning and pattern recognition capabilities than true computational competence in chemistry tasks, underscoring the need for more rigorous quantitative benchmarks.

Another challenge in current chemistry benchmarks is the lack of standardized difficulty calibration. Existing benchmarks span a wide range of problem complexities. Some focus exclusively on Olympiad-level tasks designed for high-achieving students in competitive settings~\cite{he2024olympiadbench, huang2024olympicarena}, while others include relatively simple high school or college-entry questions~\cite{hendrycks2020measuring, wang2023scibench}, which pose little challenge to experienced students such as IChO participants. A more systematic categorization of problem difficulty is essential for meaningful evaluation across ability levels. Furthermore, evaluating quantitative chemistry problems introduces a critical methodological tension in answer verification. On one hand, the objective rigor offered by strict verification systems, such as xVerify \cite{chen2025xverify}, is essential. By relying on exact matching, these tools are vital for ensuring computational soundness and penalizing answers derived from flawed reasoning. On the other hand, this deterministic approach can be fundamentally misaligned with chemical practice. Unlike pure mathematics, quantitative chemistry inherently involves numerical tolerances, significant figures, and rounding conventions derived from experimental data. A verification process that is too rigid risks unfairly penalizing models that reason correctly but yield a numerically equivalent, yet non-identical, output. It motivates a core aspect of our work: not merely to choose one method over the other, but to employ both strict and tolerance-based verification. By analyzing the performance gap between these two approaches, we can uncover deeper insights into a model's underlying reasoning and self-correction capabilities.

\noindent\textbf{Main Contributions.} 
In this work, we make the following key contributions:

\noindent(1) We design the benchmark called \textbf{QCBench}, which is specifically focused on \textbf{Q}uantitative \textbf{C}hemistry problems, such as tasks requiring explicit numerical computation, and comprising 350 problems across seven major subfields with three difficulty levels. The benchmark includes both expert-curated problems and samples adapted from existing datasets. All problems can be formalized and solved using symbolic or neural reasoning methods.

\noindent(2) We provide the first systematic analysis of their capabilities on domain-specific quantitative chemistry. Our findings reveal a critical gap between linguistic fluency and computational accuracy, a consistent performance decline with increasing problem difficulty, and distinct model-specific strengths and weaknesses across different chemical domains.

\noindent(3) We introduce a methodological insight by analyzing the performance gap between strict (\textit{e.g.}, xVerify) and tolerance-based verification. We demonstrate that for advanced models, this gap serves not as a measure of formatting error, but as a powerful indicator of sophisticated self-correction capabilities, transforming the verification process itself into a diagnostic tool for advanced reasoning.

\section{Related Works}


\begin{table*}[!htb]
\centering
\begin{adjustbox} {max width=\textwidth} \footnotesize
\begin{tabular}{@{}lc|ccccccc@{}}
\toprule
\multirow{2}{*}{\textbf{Benchmark}} & \multirow{2}{*}{\textbf{Reference}} & \multicolumn{7}{c}{\textbf{Subcategories (Only quantitative problem counted)}}  \\
\cmidrule(lr){3-9}
& & PhC & IC & QC & AC & BOC & GC & PoC  \\
\midrule
SciEval & \cite{sun2024scieval} & &  & & & & &  \\
OlympicArena & \cite{huang2024olympicarena} &\checkmark &\checkmark &  & \checkmark & &\checkmark   & \\
SciBench & \cite{wang2023scibench} & \checkmark & & \checkmark & & & &  \\
JEEBench & \cite{arora2023have} & \checkmark &\checkmark &  & & & &   \\
ChemBench & \cite{mirza2024large} & \checkmark & \checkmark & & \checkmark &  & & \checkmark  \\
OlympiadBench & \cite{he2024olympiadbench} & & & & & & & \\
ScienceQA & \cite{saikh2022scienceqa} & & & & & & & \\
SciCode & \cite{tian2024scicode} & & & \checkmark& \checkmark & & & \\
\textbf{QCBench} & Ours & \checkmark & \checkmark & \checkmark & \checkmark & \checkmark & \checkmark & \checkmark \\
\bottomrule
\end{tabular}
\end{adjustbox}
\vspace{-0.2cm}
\caption{Comparison of chemistry benchmarks by subcategory coverage. }
\label{tab:chemistry_benchmark_subcategories}
\end{table*}
\subsection{Chemistry Benchmarks}
We categorize existing chemistry benchmarks into two types based on input modality: single-modality benchmarks, which present problems purely in textual form, and multi-modality benchmarks, which incorporate both textual and visual inputs, such as molecular structures or crystallographic images. Representative single-modal benchmarks include ChemBench~\cite{mirza2024large}, ChemLLMBench~\cite{guo2023can}, MMLU~\cite{hendrycks2020measuring}, SciBench~\cite{wang2023scibench}, SciCode~\cite{tian2024scicode}, JEEBench\cite{arora2023have}, and OlympicArena\cite{he2024olympiadbench}. In contrast, multi-modal chemistry benchmarks such as ScienceQA~\cite{saikh2022scienceqa}, OlympiadBench~\cite{he2024olympiadbench}, MaCBench~\cite{alampara2024macbench}, ChemTable~\cite{zhou2025benchmarking}, MolPuzzle~\cite{guo2024can}, Emma~\cite{hao2025can}, and ScemQA~\cite{liang2024scemqa} often emphasize visual understanding and concept recognition tasks.
While multi-modal benchmarks have grown in number and scope, they are predominantly centered on conceptual or interpretive tasks, which fall outside the scope of our study. In this work, we focus on single-modality benchmarks and further analyze their coverage of chemistry subfields. Specifically, we classify benchmark questions into seven vertical domains of chemistry and summarize this categorization in Table~\ref{tab:chemistry_benchmark_subcategories}. Our analysis reveals that most existing benchmarks focus narrowly on one or two types of computational problems, lacking both domain coverage and problem diversity. This motivates the development of our benchmark, which aims to address these limitations through broader coverage and a focus on quantitative problem-solving.

\subsection{LLMs for Solving Chemistry Problems}
Recent advances in large language models (LLMs) have enabled them to tackle a broad range of chemistry problems, which can be broadly categorized into three types: qualitative, predictive, and quantitative (Figure \ref{fig:define_quantative_chemistry}). \textit{Qualitative} chemistry problems require reasoning or structural understanding, such as assigning functional groups in spectra, interpreting reaction mechanisms, or deducing morphology–property relationships. GPT-4 has shown strong performance in such tasks, although it still struggles with complex reasoning in stereochemistry or crystallography \cite{wang2023scibench, hendrycks2020measuring}. \textit{Predictive} chemistry problems involve direct property or reaction outcome predictions, where answers are derived from learned associations rather than explicit computation. These include predicting boiling points, reactivity, or products of simple reactions, often without requiring intermediate steps. Closed-source models like GPT-4 and GPT-4o have achieved notable accuracy on such tasks \cite{guo2023can}.
 Recent study has shown that o1 is the best model across calculation, knowledge inference and reasoning \cite{mirza2024large}.  Despite recent progress, most benchmarks report only aggregate accuracy, without differentiating model performance across these problem types or difficulty levels. This limits our understanding of LLM strengths and weaknesses in sub-chemistry domains, and motivates the development of more structured evaluations.
\begin{figure*}[!htb]
    \centering
    \includegraphics[width=\linewidth]{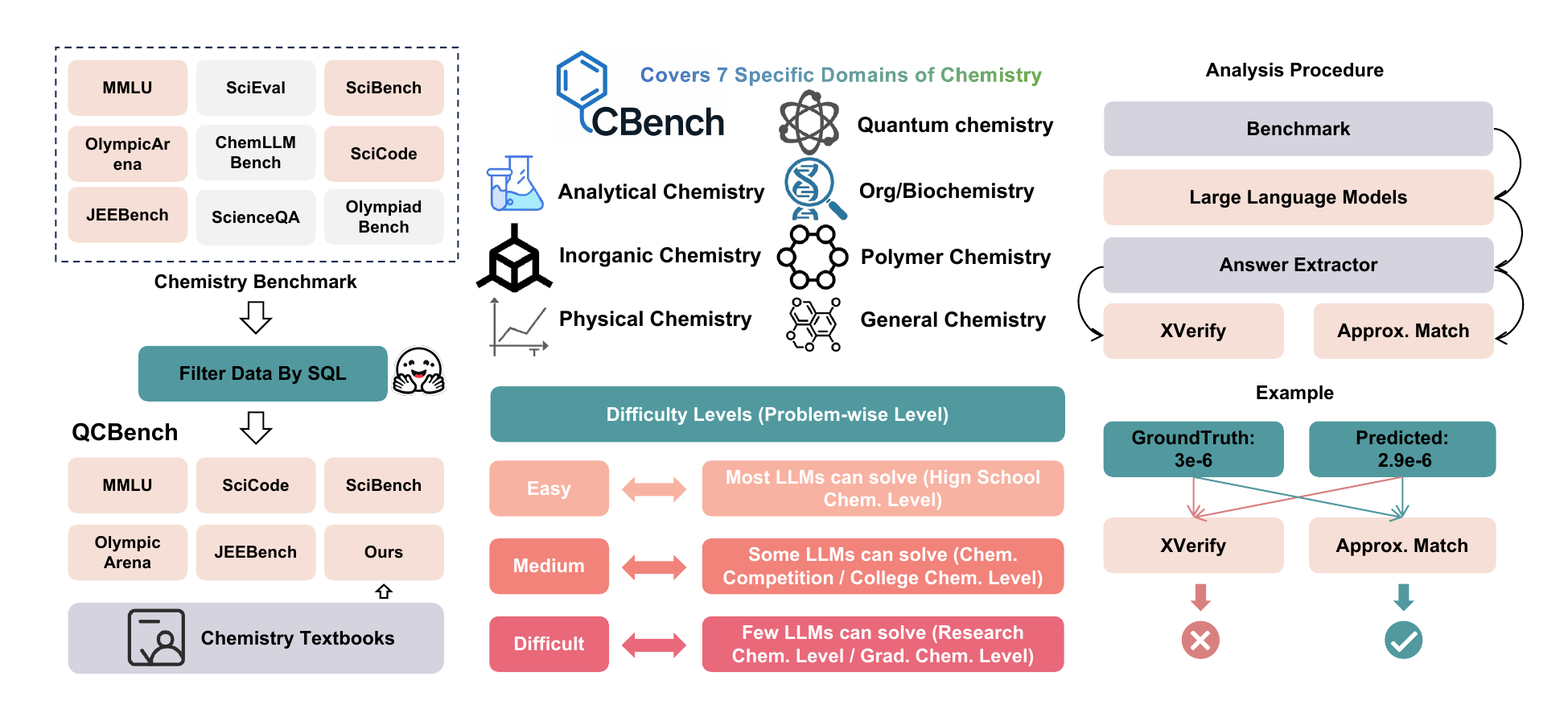}
    \vspace{-0.45cm}
    \caption{Framework of \mname. }
    \label{fig:QCBench}
\end{figure*}

\section{QCBench}

Quantitative chemistry problems require numerical reasoning grounded in formulas, constants, and multi-step derivations, such as computing Gibbs free energy or equilibrium concentrations. These tasks remain difficult for LLMs, especially in \textit{Quantum} and \textit{Physical Chemistry}, as shown by poor performance on SciBench~\cite{wang2023scibench} and college-level MMLU~\cite{hendrycks2020measuring}. To systematically assess this capability, we propose QCBench (Figure~\ref{fig:QCBench}), the \textbf{Q}uantitative \textbf{C}hemistry \textbf{Bench}mark, specifically designed for quantitative problems defined in the field of chemistry. It consists of three main components: data filtering and curation, large language model evaluation, and answer verification.

\begin{table}[!htb]
\centering
\begin{adjustbox}{max width=0.92\textwidth}\footnotesize
\begin{tabular}{lcccc}
\toprule
\textbf{Class} & \textbf{Abbrv.} & \textbf{Easy} & \textbf{Medium} & \textbf{Difficult} \\
\midrule
Analytical Chemistry      & AC    &    6 &      4 &     15 \\
Bio/Organic Chemistry     & BOC   &    2 &      9 &     14 \\
General Chemistry         & GC    &    2 &      4 &     10 \\
Inorganic Chemistry       & IC    &   11 &     16 &     19 \\
Physical Chemistry        & PhC   &   29 &     74 &     84 \\
Polymer Chemistry         & PoC   &    1 &      8 &      3 \\
Quantum Chemistry         & QC    &   19 &     13 &      7 \\
\midrule
Total                     &       &   70 &    128 &    152 \\
\bottomrule
\end{tabular}
\end{adjustbox}
\vspace{-0.2cm}
\caption{Distribution of question difficulties across chemistry subfields in full QCBench (350 questions) by question length.}
\label{tab:qcbench_difficulty}
\end{table}

\begin{table}[!htb]
\centering
\begin{adjustbox}{max width=0.92\textwidth}\footnotesize
\begin{tabular}{lcccc}
\toprule
\textbf{Class} & \textbf{Abbrev.} & \textbf{Easy} & \textbf{Medium} & \textbf{Difficult} \\
\midrule
Analytical Chemistry      & AC    &    2 &      2 &      8 \\
Bio/Organic Chemistry     & BOC   &    1 &      5 &      6 \\
General Chemistry         & GC    &    2 &      3 &      7 \\
Inorganic Chemistry       & IC    &    1 &      5 &      6 \\
Physical Chemistry        & PhC   &    1 &      3 &      8 \\
Polymer Chemistry         & PoC   &    1 &      8 &      3 \\
Quantum Chemistry         & QC    &    5 &      5 &      2 \\
\midrule
Total                     &       &   13 &     31 &     40 \\
\bottomrule
\end{tabular}
\end{adjustbox}
\vspace{-0.2cm}
\caption{Distribution of question difficulties across chemistry subfields in subset QCBench (84 questions) by question length.}
\label{tab:qcbench_balanced_difficulty}
\end{table}

\subsection{Data Preparation}

Our benchmark is constructed from two primary sources: (1) curated and annotated by human experts, and (2) collected from existing chemistry benchmarks.

\paragraph{(1) Human expert curation}
To supplement gaps in existing benchmarks, we curated additional problems through manual annotation by a chemistry Ph.D. student, with verification by senior domain experts. These problems are sourced from authoritative textbooks including \textit{Fundamentals of Analytical Chemistry} \cite{skoog1996fundamentals}, \textit{Atkins’ Physical Chemistry} \cite{atkins2023atkins}, \textit{Inorganic Chemistry} \cite{housecroft2008inorganic}, \textit{Polymer Science and Technology} \cite{ebewele2000polymer}, and \textit{Garrett’s Biochemistry} \cite{garrett2015biochemistry}. To increase coverage across domains and difficulty levels, we prioritize complex problems in underrepresented areas such as \textit{Quantum}, \textit{Analytical}, and \textit{Polymer  Chemistry}, while filtering out overly simple questions. To stratify these problems by difficulty, we employed a two-stage approach. Initially, we adopted a length-based heuristic, hypothesizing that problem difficulty correlates with input length. Based on this, we established three tiers (easy, medium, hard) using token count thresholds of 200 and 400. This initial categorization is analyzed in our results section. However, recognizing that input length is only a surface-level proxy for complexity, we also utilized a state-of-the-art large language model, Qwen3-Max~\cite{yang2025qwen3}, to perform a more semantically nuanced re-evaluation of each problem's difficulty. The full data distribution for both categorization methods is summarized in Table~\ref{tab:qcbench_difficulty}, and Table~\ref{tab:qcbench_difficulty_qwen3}.

\paragraph{(2) Collected from existing benchmarks}
To ensure consistency, we focus exclusively on single-modality benchmarks that present problems in textual form. As such, multimodal datasets like ScienceQA~\cite{saikh2022scienceqa} and OlympiadBench~\cite{he2024olympiadbench}, which require image-based understanding, are excluded. We include 250 problems from ChemBench~\cite{mirza2024large}, MMLU~\cite{hendrycks2020measuring}, SciBench~\cite{wang2023scibench}, OlympicArena~\cite{huang2024olympicarena}, and JEEBench~\cite{arora2023have}. Initial filtering is performed using SQL-based keyword search to identify calculation-related problems. Each candidate is subsequently validated by domain experts to ensure it meets our definition of quantitative chemistry—\textit{i.e.}, problems must be formalizable and yield precise floating-point answers. In other words, the FormulaVerify step (Figure~\ref{fig:define_quantative_chemistry}) is largely carried out through expert annotation.

\paragraph{Data processing}
During data consolidation, we first ensure that every problem is associated with a well-defined unit; if no unit is specified, it is treated as dimensionless. We observed that a few problems include multiple numerical answers, which pose challenges for automatic evaluation tools such as xVerify~\cite{chen2025xverify}. To address this, we decompose each multi-answer problem into $k$ independent single-answer questions when $k$ distinct answers are provided. In addition, we introduce a novel augmentation strategy: when multiple sub-questions share the same unit and are of comparable magnitude, we append a follow-up prompt  \textit{Please give the sum of the answers} to form a new, aggregated problem. This transformation encourages numerical reasoning across related values and, to our knowledge, has not been explored in prior chemistry benchmarks.

\paragraph{Data contamination and leakage check}
To check that whether our benchmark has been contaminated, which means that the data inside our benchmark are potentially being trained before, we mainly choose partial training sets from SmolInstruct \cite{yu2024llasmol}, and test sets from SciEval \cite{sun2024scieval} and ChemCoTBench \cite{li2025beyond} which could have been trained recently in some state of the art models such as Intern-S1 \cite{bai2025intern}. As the sub-fields of our benchmark are not within the scope of forward synthesis, retro-synthesis, and SMILES to IUPAC name conversions, we kept two training corpora, including molecule captioning and molecule generation, during the contamination check process for SmolInstruct. For the $N$-gram matching, the smallest number of consecutive words is 3 while the largest is 8 under our settings. The corpus contains 1511 texts from SciEval, 1495 texts from ChemCoTBench, and 2000 total texts from SMolInstruct. For the fuzzy matching, we used the package fuzzywuzzy \cite{bosker2021using} to compute similarity between the collected corpora and our benchmark.

\begin{table}[!htb]
\centering
\caption{Data contamination audit results across different fuzzy thresholds (matching under ngram\_threshold = 8)}
\label{tab:threshold_comparison}
\resizebox{\textwidth}{!}{%
\begin{tabular}{cccccp{6cm}}
\toprule
\textbf{Fuzzy Threshold} & \textbf{Contamination Rate} & \textbf{$N$-gram Avg Score} & \textbf{Fuzzy Avg Score} & \textbf{Suspected Count} & \textbf{Examples} \\
\midrule
0.5 & 7.43\% & 0.758 & 0.541 & 26 & what is the oxidation state of, and a (SciEval)\\
\midrule
0.6 & 3.71\% & 0.758 & 0.617 & 13 & the specific heat capacity of water, the ph of the solution after (SciEval) \\
\midrule
0.7 & 3.71\% & 0.758 & 0.000 & 13 & what is the oxidation state of, what is the ph of a  (SciEval)  \\
\midrule
0.8/0.9 & 3.71\% & 0.758 & 0.000 & 13 & what is the oxidation state of, what is the ph of a (SciEval)  \\
\bottomrule
\end{tabular}%
}
\end{table}

\begin{table}[!htb]
\centering
\caption{Data contamination audit results across different $N$-gram confidence (matching under ngram\_threshold = 4)}
\label{tab:threshold_comparison_ngram}
\resizebox{\textwidth}{!}{%
\begin{tabular}{cccccp{6cm}}
\toprule
\textbf{Confidence Threshold} & \textbf{Contamination Rate} & \textbf{$N$-gram Avg Score} & \textbf{Fuzzy Avg Score} & \textbf{Suspected Count} & \textbf{Examples} \\\midrule
0.5 & 54.57\% & 0.530 & 0.000 & 191 & of the solution after, calculate the number of, the molecular mass of (SciEval) \\\midrule
0.6 & 17.43\% & 0.647 & 0.000 & 61 & at the normal boiling point, the moment of inertia of (SciEval) \\\midrule
0.7 & 3.71\% & 0.758 & 0.000 & 13 & what is the oxidation state of, what is the ph of a (SciEval)  \\\midrule
0.8 & 0.29\% & 0.875 & 0.000 & 1 & the ph of the solution after the (SciEval) \\\bottomrule 
\end{tabular}%
}
\end{table}

With the $N$-gram threshold fixed at 8, we tested the impact of varying fuzzy matching thresholds from 0.5 to 0.9 on contamination detection (Table~\ref{tab:threshold_comparison}). The experimental results show a clear declining trend in contamination rate as the fuzzy threshold increases: from 7.43\% (26 suspected items) down to 3.71\% (13 suspected items). The fuzzy average score exhibits a rise-then-fall pattern: 0.541 at threshold 0.5, peaking at 0.617 at threshold 0.6, then dropping to 0.000 for thresholds 0.7 and above, suggesting that overly high fuzzy thresholds ($\geq$0.7) completely suppress fuzzy matching detection. The contamination rate is calculated as:
$$\text{Contamination Rate} = \frac{\text{Suspected Count}}{\text{\# Total Questions}} \times 100\%. $$

With the $N$-gram threshold fixed at 4, we tested the impact of confidence thresholds from 0.5 to 0.8 on exact matching detection (Table~\ref{tab:threshold_comparison_ngram}). The experimental results reveal an even more dramatic trend: the contamination rate drops precipitously from 54.57\% (191 suspected items) to just 0.29\% (only 1 suspected item), representing a reduction of over 99\%. Simultaneously, the $N$-gram average score steadily improves from 0.530 to 0.875, indicating that higher confidence thresholds effectively filter for more precise $N$-gram matches. The average score is calculated as:
$$\text{Average Score} = \frac{\sum \text{Confidence Scores}}{\text{\# Detections}}. $$
These two experiments clearly demonstrate the significant impact of different detection strategies and threshold settings on data contamination audit effectiveness, providing crucial empirical evidence to select appropriate detection parameters. Apart from these, we also observed some cases where the $N$-gram detection or fuzzy matching detection outputs some examples that are highly similar to the corpora. After the careful check, we found that these words are not actually contaminated

\paragraph{Subset QCBench-84 vs. QCBench-350} From Table \ref{tab:qcbench_difficulty}, we observe that QCBench is an imbalanced dataset, where a large proportion of problems fall into the Physical Chemistry domain. To enable a fairer comparison across domains, we curate a balanced subset, QCBench-84, from the original QCBench (QCBench-350). Specifically, we randomly sample 12 problems for each domain, with a fixed random seed to ensure reproducibility and to avoid any subjective bias in selection. While QCBench-350 serves as the full-scale benchmark for comprehensive evaluation, QCBench-84 provides a smaller and more balanced testbed. The statistics of QCBench-84 is presented in Table~\ref{tab:qcbench_balanced_difficulty} and Table~\ref{tab:qcbench_balanced_difficulty_qwen3}.

\subsection{Assessing LLMs on QCBench} As shown in Table~\ref{tab:chemistry-model-table-split} and Table~\ref{tab:chemistry-model-table-split-2}, we evaluated a total of 24 large language models on QCBench, including 11 proprietary models, such as Claude 3.5 Sonnet, GPT-4o \cite{hurst2024gpt}, and Gemini 2.5 pro \cite{comanici2025gemini} and 13 open-source models, such as DeepSeek-R1 \cite{guo2025deepseek}, and Qwen3-235B \cite{yang2025qwen3}. These models were selected to cover a broad spectrum of capabilities, licensing terms, and training scales, ensuring a comprehensive assessment of current LLM performance in quantitative chemistry tasks.
\subsection{Answer Verification}
We adopt two complementary approaches for answer verification: (i) a strict verifier based on xVerify, and (ii) a custom tolerance-based pipeline tailored for chemistry problems. 

\noindent\textbf{(i) xVerify-based strict verifier}. 
For tasks with deterministic answers, common in mathematics and physics, xVerify~\cite{chen2025xverify} serves as a reliable tool. Specifically, we use the scalable version of xVerify-0.5B-I to evaluate the output of each LLM.

\noindent\textbf{(ii) Tolerance-based evaluation pipeline.} 
Unlike mathematics and physics, in quantitative chemistry, slight approximations are often acceptable due to the nature of experimental data and rounding conventions. To accommodate this, we design a relaxed evaluation pipeline that tolerates small relative errors while maintaining fairness. Specifically, we provide two verification modes: 
\begin{itemize}[leftmargin=*]
\item Basic mode: Supports exact numeric formats including integers, floating-point numbers, scientific notation (\textit{e.g.}, $a \times 10^{-b}$, $a*e{-b}$), fractions ($\frac{a}{b}$), and mixed fractions ($5\frac{a}{b}$). A default relative error tolerance of $1\times10^{-6}$ is applied.
\item Pro mode: Incorporates decimal precision matching by comparing the number of decimal places in the predicted answer with that of the ground truth, allowing for context-aware flexibility. 
\end{itemize}
For fair comparison, we adopt the pro mode by default. The ablation study on tolerance error under the basic mode is discussed later.

\section{Experiment Setup}

\paragraph{Prompt}
The system prompt is "You are an expert chemist. Please read the following question and provide a step-by-step solution. Your final answer must be presented as a readable LaTeX formula, enclosed in a \text{\\\boxed\{\\\boxed\}} environment. If the final answer is numerical, write only the numeric value inside \text{\text{\\\boxed\{\\\boxed\}}}; place the unit immediately after the box (not inside), using the unit specified in the problem." For the user prompt, if there is a specific unit associated with problem, we add the \textit{The unit of the final answer is \text{\\\boxed\{\\\boxed\}}} after the question description, otherwise we only append the text \textit{Do not put the unit inside the \text{\\\boxed\{\\\boxed\}}; place it right after the box.} to the end of the question.

\paragraph{Hyper-parameters}
Model inference was conducted via the OpenRouter API, a public LLM service provider. To ensure comparability, we applied a standardized decoding strategy across models, setting the temperature to 0.1 for near-deterministic outputs and top\_p to 1.0. The default maximum output tokens was set to 16,384. This limit was increased to 65,535 for models known for verbose chain-of-thought reasoning, such as DeepSeek-R1 and Gemini-2.5-pro, to prevent premature truncation of their outputs. Due to the constraints of using external APIs, we could not set a fixed random seed, and we refrained from defining custom stop words to avoid conflicts with providers' internal configurations. To handle API failures, each request was configured with up to three retry attempts. The full list of evaluated models is provided in Table~\ref{tab:chemistry-model-table-split} and~\ref{tab:chemistry-model-table-split-2}.

\begin{sidewaystable*}[!htbp]
  \centering
  \footnotesize 
  \setlength\tabcolsep{1.9pt}
  \renewcommand{\arraystretch}{0.92}
  \begin{tabular}{l| ll|ll|ll|ll|ll|ll|ll|ll}
    \toprule
    \multirow{2}{*}{\textbf{Model}} & \multicolumn{2}{c|}{\textbf{Analytical}} & \multicolumn{2}{c|}{\textbf{Biochemistry}} & \multicolumn{2}{c|}{\textbf{General}} & \multicolumn{2}{c|}{\textbf{Inorganic}} & \multicolumn{2}{c|}{\textbf{Physical}} & \multicolumn{2}{c|}{\textbf{Polymer}} & \multicolumn{2}{c|}{\textbf{Quantum}} & \multicolumn{2}{c}{\textbf{Avg. (Perf.)}} \\
    \cmidrule(lr){2-3} \cmidrule(lr){4-5} \cmidrule(lr){6-7} \cmidrule(lr){8-9} \cmidrule(lr){10-11} \cmidrule(lr){12-13} \cmidrule(lr){14-15} \cmidrule(lr){16-17}
    & Appr. & xVerify & Appr. & xVerify & Appr. & xVerify & Appr. & xVerify & Appr. & xVerify & Appr. & xVerify & Appr. & xVerify & Appr. & xVerify \\
    \midrule
    \rowcolor{gray!20} \multicolumn{17}{c}{\textit{Closed-source Models}} \\
    Claude-3.5-Sonnet & $19.4_{4.8}$ & $13.9_{9.6}$ & $30.6_{4.8}$ & $33.3_{0.0}$ & $27.8_{4.8}$ &  $27.8_{4.8}$ &  $50.0_{0.0}$ &  $50.0_{0.0}$ & $33.3_{0.0}$ & $33.3_{0.0}$ & $\underline{33.3_{14.4}}$ & $27.8_{9.6}$ & $25.0_{8.3}$ & $22.2_{9.6}$ & $31.3_{4.5}$ & $29.2_{3.7}$\\
    Claude-4-Sonnet & $37.5_{5.9}$ & $37.5_{5.9}$ & $\underline{41.7_{0.0}}$ & $\underline{41.7_{0.0}}$ & $45.8_{5.9}$ & $\underline{54.2_{5.9}}$ & $58.3_{11.8}$ & $58.3_{11.8}$ & $\mathbf{41.7_{0.0}}$ & $37.5_{5.9}$ & $\underline{33.3_{0.0}}$ & $29.2_{5.9}$ & $29.2_{5.9}$ & $29.2_{5.9}$ & $41.1_{0.8}$ & $40.7_{0.9}$ \\
    Gemini-2.5-pro & $38.9_{4.8}$ & $41.7_{0.0}$ & $30.6_{4.8}$ & $38.9_{9.6}$ & $\underline{47.2_{4.8}}$ & $47.2_{4.8}$ & $72.2_{12.7}$ & $75.0_{8.3}$ & $19.4_{4.8}$ & $19.4_{4.8}$ & $27.8_{4.8}$ & $27.8_{4.8}$ & $41.7_{8.3}$ & $41.7_{8.3}$ & $39.7_{3.8}$ & $41.8_{3.1}$ \\
    Gemini-2.5-flash & $36.1_{17.3}$ & $41.7_{22.0}$ & $25.0_{8.3}$ & $27.8_{4.8}$ & $33.3_{0.0}$ & $33.3_{0.0}$ & $\underline{75.0_{8.3}}$ & $72.2_{4.8}$ & $30.6_{4.8}$ & $27.8_{4.8}$ & $22.2_{9.6}$ & $19.4_{4.8}$ & $38.9_{19.2}$ & $36.1_{17.3}$ & $37.3_{2.5}$ & $37.7_{3.6}$\\
    GPT-4o  & $19.4_{4.8}$ & $13.9_{4.8}$ & $22.2_{4.8}$  & $22.2_{4.8}$ & $25.0_{8.3}$ & $27.8_{4.8}$  & $52.8_{9.6}$ & $52.8_{9.6}$ & $11.1_{4.8}$ & $11.1_{4.8}$ & $22.2_{4.8}$ & $13.9_{4.8}$ & $13.9_{4.8}$ & $13.9_{4.8}$  & $23.8_{1.2}$ & $22.3_{1.4}$ \\
    Grok-3 & $44.4_{4.8}$ & $41.7_{8.3}$ & $27.8_{4.8}$ & $33.3_{8.3}$ & $30.6_{9.6}$ & $33.3_{8.3}$ & $63.9_{4.8}$ & $66.7_{8.3}$ & $\underline{38.9_{4.8}}$ & $\underline{38.9_{4.8}}$ & $19.4_{4.8}$ & $25.0_{8.3}$ & $47.2_{4.8}$ & $47.2_{4.8}$ & $38.9_{1.8}$ & $41.9_{2.6}$ \\
    Grok-4 & $41.7_{0.0}$ & $\mathbf{50.0_{0.0}}$  & $25.0_{0.0}$ & $33.3_{0.0}$ & $\mathbf{50.0_{0.0}}$ & $50.0_{0.0}$ & $75.0_{0.0}$ & $75.0_{0.0}$ & $37.5_{5.9}$ & $33.3_{0.0}$ & $25.0_{0.0}$ & $25.0_{0.0}$ & $\mathbf{62.5_{5.9}}$ & $\mathbf{62.5_{5.9}}$ & $\underline{45.2_{0.0}}$ & $\underline{48.3_{0.9}}$\\
    o3 & $\underline{50.0_{0.0}}$ & $\mathbf{50.0_{0.0}}$ & $33.3_{8.3}$ & $41.7_{8.3}$ & $44.4_{4.8}$ & $50.0_{8.3}$ & $72.2_{4.8}$ & $72.2_{4.8}$ & $\mathbf{41.7_{8.3}}$ & $\mathbf{41.7_{8.3}}$ & $\underline{33.3_{8.3}}$ & $27.8_{4.8}$ & $\underline{61.1_{4.8}}$ & $\underline{61.1_{4.8}}$ & $\mathbf{48.0_{1.8}}$ & $\mathbf{49.7_{0.7}}$\\
    o3-mini & $41.7_{8.3}$ & $41.7_{8.3}$ & $25.0_{0.0}$ & $36.1_{4.8}$ & $25.0_{8.3}$ & $25.0_{8.3}$ & $\mathbf{80.6_{4.8}}$ & $\mathbf{80.6_{4.8}}$ & $33.3_{0.0}$ & $33.3_{0.0}$ & $22.2_{4.8}$ & $19.4_{4.8}$ & $38.9_{9.6}$ & $38.9_{9.6}$ & $38.1_{2.4}$ & $39.4_{2.0}$ \\
    o4-mini & $47.2_{4.8}$ & $\mathbf{50.0_{0.0}}$ & $33.3_{0.0}$ & $41.7_{0.0}$ & $36.1_{4.8}$ & $36.1_{4.8}$ & $72.2_{12.7}$ & $72.2_{12.7}$ & $33.3_{8.3}$ & $\mathbf{41.7_{8.3}}$ & $\underline{33.3_{8.3}}$ & $27.8_{9.6}$ & $47.2_{12.7}$ & $47.2_{12.7}$ & $43.3_{3.4}$ & $46.4_{2.2}$\\
    GPT-5-mini &  $41.7_{8.3}$ &  $41.7_{0.0}$ & $19.4_{4.8}$ & $22.2_{9.6}$ & $36.1_{12.7}$ & $36.1_{12.7}$ & $72.2_{4.8}$ & $77.8_{4.8}$ & $19.4_{9.6}$ & $27.8_{9.6}$ & $30.6_{4.8}$ & $\mathbf{36.1_{4.8}}$ & $41.7_{8.3}$ & $41.7_{8.3}$ & $37.3_{2.5}$ & $41.3_{2.5}$\\

    \midrule
    \rowcolor{gray!20} \multicolumn{17}{c}{\textit{Open-source Models}} \\
    Qwen2.5-72B & $13.9_{4.8}$ & $8.3_{8.3}$ & $16.7_{0.0}$ & $16.7_{0.0}$ & $16.7_{0.0}$ & $19.4_{4.8}$ & $30.6_{9.6}$ & $30.6_{9.6}$ & $13.9_{4.8}$ & $11.1_{4.8}$ & $25.0_{0.0}$ & $16.7_{0.0}$ & $16.7_{0.0}$ & $16.7_{0.0}$ & $19.0_{2.4}$ & $17.5_{3.0}$\\
    Qwen3-32B & $19.4_{12.7}$ & $16.7_{14.4}$ & $33.3_{8.3}$ & $36.1_{4.8}$ & $25.0_{8.3}$ & $38.9_{12.7}$ & $52.8_{4.8}$ & $52.8_{4.8}$ & $27.8_{4.8}$ & $30.6_{4.8}$ & $22.2_{4.8}$ & $22.2_{4.8}$ & $27.8_{4.8}$ & $27.8_{4.8}$ & $29.8_{0.0}$ & $31.7_{1.2}$\\
    Qwen3-235B & $27.8_{4.8}$ & $27.8_{4.8}$ & $27.8_{4.8}$ & $30.6_{4.8}$ & $25.0_{0.0}$ & $27.8_{4.8}$ & $50.0_{0.0}$ & $50.0_{0.0}$ & $36.1_{9.6}$ & $36.1_{9.6}$ & $30.6_{4.8}$ & $30.6_{4.8}$ & $44.4_{4.8}$ & $47.2_{4.8}$ & $34.5_{2.4}$ & $35.9_{2.5}$\\
    QwQ-32B & $45.8_{5.9}$  & $\underline{45.8_{5.9}}$ & $25.0_{0.0}$ & $29.2_{5.9}$ & $37.5_{5.9}$ & $37.5_{5.9}$ & $\underline{75.0_{0.0}}$ & $\underline{79.2_{5.9}}$ & $25.0_{0.0}$ & $25.0_{0.0}$ & $\mathbf{37.5_{5.9}}$ & $\underline{33.3_{0.0}}$ & $50.0_{0.0}$ & $50.0_{0.0}$ & $42.3_{0.8}$ & $42.6_{1.8}$\\
    DeepSeek-V3 & $33.3_{16.7}$ & $27.8_{17.3}$ & $36.1_{4.8}$ & $36.1_{4.8}$ & $33.3_{8.3}$ & $36.1_{9.6}$ & $55.6_{4.8}$ & $55.6_{4.8}$ & $36.1_{17.3}$ & $36.1_{17.3}$ & $27.8_{4.8}$ & $22.2_{9.6}$ & $36.1_{4.8}$ & $36.1_{4.8}$ & $36.9_{2.1}$ & $35.4_{1.1}$\\
    DeepSeek-V3.1 & $\mathbf{52.8_{9.6}}$ & $44.4_{9.6}$ & $30.6_{4.8}$ & $30.6_{4.8}$ & $27.8_{4.8}$ & $30.6_{12.7}$ & $58.3_{14.4}$ & $58.3_{14.4}$ & $27.8_{4.8}$ & $27.8_{4.8}$ & $30.6_{4.8}$ & $22.2_{4.8}$ & $44.4_{4.8}$ & $38.9_{9.6}$ & $38.9_{2.5}$ & $36.1_{3.8}$ \\
    DeepSeek-R1  & $41.7_{11.8}$ & $\mathbf{50.0_{0.0}}$ & $25.0_{0.0}$ & $37.5_{5.9}$ & $33.3_{23.6}$ & $41.7_{11.8}$ & $62.5_{5.9}$ & $62.5_{5.9}$ & $33.3_{0.0}$ & $33.3_{0.0}$ & $20.8_{17.7}$ & $20.8_{17.7}$ & $61.0_{3.7}$ & $61.0_{3.7}$ & $39.7_{7.3}$ & $45.2_{3.1}$ \\
    Gemma-3-27B-it & $16.7_{0.0}$ & $8.3_{0.0}$ & $27.8_{4.8}$ & $27.8_{4.8}$ & $19.4_{4.8}$ &$19.4_{4.8}$ & $44.4_{4.8}$ & $44.4_{4.8}$ & $16.7_{14.4}$ & $16.7_{14.4}$ & $13.9_{4.8}$ & $11.1_{4.8}$ & $36.1_{4.8}$ & $36.1_{4.8}$ & $25.0_{1.2}$ & $24.6_{0.1}$\\
    Llama-3-405b-I & $13.9_{4.8}$  & $5.6_{4.8}$ & $25.0_{8.3}$ & $25.0_{8.3}$ & $19.4_{15.7}$ & $22.2_{12.7}$ & $30.6_{4.8}$ & $30.6_{4.8}$ & $13.9_{4.8}$ & $11.1_{4.8}$ & $19.4_{4.8}$ & $11.1_{4.8}$ & $16.7_{8.3}$ & $16.7_{8.3}$ & $19.8_{2.7}$ & $17.1_{3.3}$\\
    Llama-3.3-70b-I  & $8.3_{0.0}$ & $0.0_{0.0}$ & $16.7_{8.3}$ & $16.7_{8.3}$ & $22.2_{9.6}$ & $16.7_{14.4}$ & $27.8_{9.6}$ & $30.6_{4.8}$ & $16.7_{0.0}$ & $16.7_{0.0}$ & $25.0_{8.3}$ & $16.7_{8.3}$ & $33.3_{8.3}$ & $36.1_{4.8}$ & $21.4_{2.4}$ & $18.9_{3.0}$ \\
    GPT-oss-20B & $\underline{50.0_{0.0}}$ & $\mathbf{50.0_{0.0}}$ & $25.0_{0.0}$ & $38.9_{4.8}$ & $\mathbf{50.0_{8.3}}$ & $\mathbf{61.1_{9.6}}$ & $61.1_{4.8}$ & $61.1_{4.8}$ & $25.0_{8.3}$ & $33.3_{8.3}$ & $\underline{33.3_{0.0}}$ & $27.8_{4.8}$ & $36.1_{9.6}$ & $36.1_{9.6}$ & $40.1_{2.7}$ & $44.2_{2.6}$ \\
    GPT-oss-120B & $\mathbf{52.8_{4.8}}$ & $\mathbf{50.0_{0.0}}$ & $\mathbf{47.2_{9.6}}$ & $\mathbf{58.3_{8.3}}$ & $33.3_{8.3}$ & $44.4_{4.8}$ & $55.6_{21.0}$ & $66.7_{16.7}$ & $19.4_{4.8}$ & $25.0_{0.0}$ & $\underline{33.3_{0.0}}$ & $25.0_{0.0}$ & $38.9_{12.7}$ & $38.9_{12.7}$ & $40.1_{4.8}$ & $44.4_{3.7}$\\
    Llama-4-Scout & $27.8_{4.8}$ & $19.4_{4.8}$ & $36.1_{4.8}$ & $38.9_{4.8}$ & $33.3_{8.3}$ & $36.1_{4.8}$ & $55.6_{9.6}$ & $55.6_{9.6}$ & $36.1_{9.6}$ & $30.6_{4.8}$ & $30.6_{4.8}$ & $25.0_{0.0}$ & $52.8_{4.8}$ & $52.8_{4.8}$ & $38.9_{5.0}$ & $36.4_{2.5}$\\
    
    \midrule
    \rowcolor{gray!10} \textbf{Avg. (Task)}  & $33.8_{15.2}$ & $31.6_{18.2}$ & $28.6_{8.4}$ & $32.0_{10.4}$ & $31.9_{11.5}$ & $34.9_{13.2}$ & $58.0_{16.4}$ & $59.1_{16.4}$ & $27.5_{11.2}$ & $28.2_{10.8}$ & $27.1_{7.9}$ & $23.3_{8.2}$ & $38.6_{14.5}$ & $41.0_{15.7}$  & - & -\\
    \bottomrule
  \end{tabular}
  \vspace{-0.2cm}

  \caption{Approximation and Exact Matching (by xVerify) Accuracies (\%, $\uparrow$) of models across 7 chemistry subfields, with columns for Approximate (Appr.) and xVerify results. \textbf{Bold} indicates the best score per column, \underline{underlined} indicates the second-best (excluding ties). Our experiments employ xVerify-0.5B-I. This is the result for the balanced benchmark dataset, with 95\% confidence interval}
  \label{tab:chemistry-model-table-split}
\end{sidewaystable*}

\begin{sidewaystable*}[!htbp]
  \centering
  \footnotesize  
  \setlength\tabcolsep{1.9pt}
  \renewcommand{\arraystretch}{0.92}
  \begin{tabular}{l| ll|ll|ll|ll|ll|ll|ll|ll}
 
    \toprule
    \multirow{2}{*}{\textbf{Model}} & \multicolumn{2}{c|}{\textbf{Analytical}} & \multicolumn{2}{c|}{\textbf{Biochemistry}} & \multicolumn{2}{c|}{\textbf{General}} & \multicolumn{2}{c|}{\textbf{Inorganic}} & \multicolumn{2}{c|}{\textbf{Physical}} & \multicolumn{2}{c|}{\textbf{Polymer}} & \multicolumn{2}{c|}{\textbf{Quantum}} & \multicolumn{2}{c}{\textbf{Avg. (Perf.)}} \\
    \cmidrule(lr){2-3} \cmidrule(lr){4-5} \cmidrule(lr){6-7} \cmidrule(lr){8-9} \cmidrule(lr){10-11} \cmidrule(lr){12-13} \cmidrule(lr){14-15} \cmidrule(lr){16-17}
    & Appr. & xVerify & Appr. & xVerify & Appr. & xVerify & Appr. & xVerify & Appr. & xVerify & Appr. & xVerify & Appr. & xVerify & Appr. & xVerify \\
    \midrule
    \rowcolor{gray!20} \multicolumn{17}{c}{\textit{Closed-source Models}} \\
    Claude-3.5-Sonnet & $18.7_{2.3}$ & $22.7_{4.6}$ & $22.7_{6.1}$ & $24.0_{4.0}$ & $20.8_{3.6}$ & $20.8_{3.6}$ & $42.0_{4.5}$ & $44.2_{3.3}$ & $31.4_{0.8}$ & $29.4_{0.5}$ & $\underline{33.3_{14.4}}$ & $27.8_{9.6}$ & $23.1_{6.8}$ & $24.8_{5.9}$ & $27.4_{4.0}$ & $27.7_{3.5}$\\
    Claude-4-Sonnet & $26.0_{2.8}$ & $32.0_{5.7}$ & $42.0_{2.8}$ & $42.0_{2.8}$ & $46.9_{4.4}$ & $\underline{53.1_{4.4}}$ & $55.4_{1.5}$ & $53.3_{1.5}$ & $41.4_{0.4}$ & $38.5_{0.0}$ & $\underline{33.3_{0.0}}$ & $29.2_{5.9}$ & $38.5_{0.0}$ & $38.5_{0.0}$ & $40.5_{0.4}$ & $40.9_{0.8}$ \\
    Gemini-2.5-pro & $30.7_{2.3}$ & $36.0_{0.0}$ & $30.7_{4.6}$ & $33.3_{2.3}$ & $\underline{47.9_{3.6}}$ & $47.9_{3.6}$ & $55.8_{3.3}$ & $58.0_{3.3}$ & $\underline{50.6_{1.3}}$ & $\underline{46.2_{2.2}}$ &  $27.8_{4.8}$ & $27.8_{4.8}$ & $42.7_{3.0}$ & $43.6_{2.6}$ & $40.9_{2.9}$ & $41.8_{1.9}$\\
    Gemini-2.5-flash & $25.3_{12.2}$ & $25.3_{12.2}$ & $33.3_{2.3}$ & $32.0_{4.0}$ & $43.8_{0.0}$ & $43.8_{0.0}$ & $57.2_{3.3}$ & $54.3_{2.2}$ & $33.0_{1.7}$ & $30.5_{0.9}$ & $22.2_{9.6}$ & $19.4_{4.8}$ & $30.8_{8.9}$ & $30.8_{6.8}$ & $35.1_{1.7}$ & $33.7_{1.6}$\\
    GPT-4o  & $16.0_{6.9}$ & $13.3_{6.1}$ & $22.7_{2.3}$ & $24.0_{0.0}$ & $20.8_{70.2}$ & $22.9_{3.6}$ & $48.6_{1.3}$ & $44.2_{2.5}$ & $22.8_{2.6}$ & $21.4_{1.9}$ & $22.2_{4.8}$ & $13.9_{4.8}$ & $23.9_{3.9}$ & $24.8_{3.0}$ & $25.3_{1.4}$ & $23.5_{1.4}$\\
    Grok-3 & $33.3_{2.3}$ & $36.0_{4.0}$ & $30.7_{6.1}$ & $37.3_{4.6}$ & $41.7_{7.2}$ & $43.8_{6.2}$ & $59.4_{4.5}$ & $60.0_{4.3}$ & $46.0_{3.5}$ & $\underline{46.2_{3.8}}$ & $19.4_{4.8}$ & $25.0_{8.3}$ & $43.6_{5.1}$ & $47.0_{3.9}$ & $39.2_{2.6}$ & $42.3_{3.1}$ \\
    Grok-4 & $28.0_{5.7}$ & $\mathbf{46.0_{2.8}}$ & $36.0_{5.7}$ & $40.0_{5.7}$ & $\mathbf{50.0_{0.0}}$ & $50.0_{0.0}$ & $63.0_{3.1}$ & $62.0_{1.5}$ & $\mathbf{52.7_{1.9}}$ &$\mathbf{51.2_{2.3}}$ & $25.0_{0.0}$ & $25.0_{0.0}$ & $48.7_{3.6}$ & $48.7_{3.6}$ & $\underline{43.3_{0.7}}$ & $\mathbf{46.1_{0.8}}$\\
    o3 & $\underline{38.7_{2.3}}$ & $33.3_{2.3}$ & $\underline{42.7_{2.3}}$ & $\underline{45.3_{4.6}}$ & $\mathbf{50.0_{0.0}}$ & $\mathbf{54.2_{3.6}}$ & $\mathbf{64.5_{1.3}}$ & $\underline{63.0_{2.2}}$ & $47.2_{3.4}$ & $45.1_{3.9}$ & $\underline{33.3_{8.3}}$ & $27.8_{4.8}$ & $\underline{51.3_{2.6}}$ & $\underline{51.3_{2.6}}$ & $\mathbf{46.8_{1.8}}$  & $\underline{45.7_{1.0}}$\\
    o3-mini & $33.3_{6.1}$ & $29.3_{6.1}$ & $33.3_{4.6}$ & $42.7_{6.1}$ & $22.9_{9.5}$ & $22.9_{9.5}$ & $\underline{63.8_{1.3}}$ & $\mathbf{63.8_{2.5}}$ & $39.8_{1.7}$ & $38.1_{1.6}$ & $22.2_{4.8}$ & $19.4_{4.8}$ & $43.6_{5.1}$ & $44.4_{6.5}$ & $37.0_{2.6}$ & $37.2_{1.9}$\\
    o4-mini & $34.7_{6.1}$ & $30.7_{2.3}$ & $37.3_{4.6}$ & $42.7_{6.1}$ & $35.4_{3.6}$ & $35.4_{3.6}$ & $\mathbf{64.5_{4.5}}$ & $\mathbf{63.8_{7.0}}$ & $40.5_{0.6}$ & $39.4_{0.8}$ & $\underline{33.3_{8.3}}$ & $27.8_{9.6}$ & $41.9_{3.9}$ & $44.4_{3.9}$ & $41.1_{2.8}$ & $40.6_{2.3}$\\
    GPT-5-mini & $30.7_{2.3}$ & $29.3_{2.3}$ & $25.3_{6.1}$ & $30.7_{8.3}$ & $33.3_{9.5}$ & $33.3_{9.5}$ & $58.7_{2.2}$ & $60.1_{1.3}$ & $39.8_{2.2}$ & $39.9_{2.0}$ & $30.6_{4.8}$ & $\mathbf{36.1_{4.8}}$ & $29.9_{3.9}$ & $34.2_{3.0}$ & $35.5_{2.4}$ & $37.7_{3.4}$\\

    \midrule
    \rowcolor{gray!20} \multicolumn{17}{c}{\textit{Open-source Models}} \\
    Qwen2.5-72B & $13.3_{4.6}$ & $6.7_{6.1}$ & $18.7_{2.3}$ & $18.7_{2.3}$ & $12.5_{0.0}$ & $14.6_{3.6}$ & $37.0_{3.8}$ & $33.3_{2.5}$ & $21.2_{0.3}$ & $19.4_{0.3}$ & $25.0_{0.0}$ & $16.7_{0.0}$ & $22.2_{3.0}$ & $21.4_{1.5}$ & $21.4_{1.2}$ & $18.7_{1.5}$ \\
    Qwen3-32B & $16.8_{8.0}$ & $13.3_{6.1}$ & $30.7_{2.3}$ & $32.0_{4.0}$ & $22.9_{3.6}$ & $33.3_{7.2}$ & $47.1_{1.3}$ & $46.4_{2.5}$ & $28.9_{2.8}$ & $28.3_{1.6}$ & $22.2_{4.8}$ & $22.2_{4.8}$ & $27.4_{3.0}$ & $27.4_{3.0}$ & $27.9_{1.8}$ & $29.0_{1.4}$\\
    Qwen3-235B & $18.7_{2.3}$ & $18.7_{2.3}$ & $29.3_{6.1}$ & $30.7_{6.1}$ & $25.0_{0.0}$ & $27.1_{3.6}$ & $50.7_{4.5}$ & $46.4_{4.5}$ & $35.8_{1.6}$ & $33.2_{1.4}$ & $30.6_{4.8}$ & $30.6_{4.8}$ & $32.5_{1.5}$ & $35.0_{3.9}$ & $31.8_{1.0}$ & $31.6_{0.9}$\\
    QwQ-32B & $32.0_{0.0}$ & $\underline{40.0_{0.0}}$ & $28.0_{0.0}$ & $30.0_{2.8}$ & $37.5_{8.8}$ & $37.5_{8.8}$ & $58.7_{3.1}$ & $58.7_{6.1}$ & $46.0_{0.8}$ & $44.1_{1.1}$ & $\mathbf{37.5_{5.9}}$ & $\underline{33.3_{0.0}}$ & $48.7_{3.6}$ & $48.7_{3.6}$ & $41.2_{0.4}$ & $41.8_{0.4}$ \\
    DeepSeek-V3 & $24.0_{6.9}$ & $24.0_{10.6}$ & $29.3_{4.6}$ & $29.3_{4.6}$ & $29.2_{7.2}$ & $31.2_{6.2}$ & $49.3_{5.0}$ & $44.9_{5.0}$ & $33.7_{1.1}$ & $31.2_{0.3}$ & $27.8_{4.8}$ & $22.2_{9.6}$ & $31.6_{1.5}$ & $31.6_{1.5}$ & $32.1_{0.9}$ & $30.7_{1.9}$\\
    DeepSeek-V3.1 & $36.0_{6.9}$ & $34.7_{4.6}$ & $33.3_{4.6}$ & $34.7_{6.1}$ & $33.3_{7.2}$ & $35.4_{7.2}$ & $52.9_{7.0}$ & $51.4_{8.2}$ & $36.2_{0.6}$ & $32.8_{1.1}$ & $30.6_{4.8}$ & $22.2_{4.8}$ & $35.0_{5.3}$ & $33.3_{6.8}$ & $36.8_{1.8}$ & $34.9_{1.6}$\\
    DeepSeek-R1 & $34.0_{8.5}$ & $46.0_{2.8}$ & $30.0_{2.8}$ & $36.0_{11.3}$ & $34.4_{13.3}$ & $43.8_{0.0}$ & $56.5_{6.1}$ & $58.7_{0.0}$ & $40.6_{1.5}$ & $39.8_{1.9}$ & $20.8_{17.7}$ & $20.8_{17.7}$ & $\mathbf{52.6_{1.8}}$ & $\mathbf{53.8_{3.6}}$ & $38.4_{6.1}$ & $42.7_{0.5}$\\
    Gemma-3-27B-it & $12.0_{0.0}$ & $4.0_{0.0}$ & $17.3_{2.3}$ & $17.3_{2.3}$ & $14.6_{3.6}$ & $14.6_{3.6}$ & $34.1_{2.5}$ & $31.9_{1.3}$ & $20.5_{1.6}$ & $20.3_{1.4}$ & $13.9_{4.8}$ & $11.1_{4.8}$ & $22.2_{7.4}$ & $26.5_{8.2}$ & $19.2_{1.0}$ & $18.0_{0.8}$\\
    Llama-3-405b-I & $13.3_{6.1}$ & ${8.0_{4.0}}$ & $17.3_{2.3}$ & $17.3_{2.3}$ & $14.6_{9.5}$ & $16.7_{9.5}$ & $26.1_{5.8}$ & $27.5_{4.5}$ & $18.7_{3.2}$ & $18.9_{2.5}$ & $19.4_{4.8}$ & $11.1_{4.8}$ & $22.2_{1.5}$ & $23.9_{3.0}$ & $18.8_{2.8}$ & $17.8_{3.3}$\\
    Llama-3.3-70b-I  & $10.7_{2.3}$ & $4.0_{0.0}$ & $16.0_{4.0}$ & $16.0_{4.0}$ & $16.7_{7.2}$ & $12.5_{10.8}$ & $31.2_{4.5}$ & $29.7_{3.3}$ & $20.7_{1.7}$ & $19.6_{2.2}$ & $25.0_{8.3}$ & $16.7_{8.3}$ & $21.4_{3.0}$ & $23.9_{3.9}$ & $20.2_{1.7}$ & $17.5_{2.3}$\\
    GPT-oss-20B & $32.0_{4.0}$ & $\underline{40.0_{0.0}}$ & $29.3_{2.3}$ & $38.7_{4.6}$ & $41.7_{9.5}$ & $50.0_{10.8}$ & $56.5_{2.2}$ & $58.0_{1.3}$ & $37.4_{3.2}$ & $38.0_{4.0}$ & $\underline{33.3_{0.0}}$ & $27.8_{4.8}$ & $37.6_{1.5}$ & $40.2_{1.5}$ & $38.3_{1.7}$ & $41.8_{1.3}$\\
    GPT-oss-120B & $\mathbf{40.0_{0.0}}$ & $32.0_{0.0}$ & $\mathbf{54.7_{6.1}}$ & $\mathbf{64.0_{0.0}}$ & $25.0_{6.2}$ & $33.3_{3.6}$ & $54.3_{7.8}$ & $59.4_{4.5}$ & $35.1_{2.2}$ & $39.4_{1.6}$ & $\underline{33.3_{0.0}}$ & $25.0_{0.0}$ & $38.5_{7.7}$ & $40.2_{7.8}$ & $40.1_{2.2}$ & $41.9_{1.8}$\\
    Llama-4-Scout & $16.0_{4.0}$ & $10.7_{2.3}$ & $29.3_{2.3}$ & $30.7_{2.3}$ & $25.0_{6.2}$ & $27.1_{3.6}$ & $48.6_{3.3}$ & $45.7_{5.8}$ & $34.2_{1.4}$ & $31.0_{0.9}$ & $30.6_{4.8}$ & $25.0_{0.0}$ & $46.2_{0.0}$ & $45.3_{1.5}$ & $32.8_{2.0}$ & $30.8_{0.6}$ \\
    \midrule
    \rowcolor{gray!10} \textbf{Avg. (Task)}  & $25.3_{10.3}$ & $24.8_{13.2}$ & $29.8_{9.5}$ & $32.7_{11.4}$ & $30.4_{12.8}$ & $32.8_{13.6}$ & $51.1_{11.2}$ & $50.4_{11.8}$ & $35.0_{9.6}$ & $33.7_{9.4}$ & $27.1_{7.9}$ & $23.3_{8.2}$ & $35.0_{10.5}$ & $36.2_{10.3}$ & - & -\\
    \bottomrule
  \end{tabular}
  \vspace{-0.2cm}

  \caption{Approximation and Exact Matching (by xVerify) Accuracies (\%, $\uparrow$) of models across 7 chemistry subfields, with columns for Approximate (Appr.) and xVerify results. \textbf{Bold} indicates the best score per column, \underline{underlined} indicates the second-best (excluding ties). Our experiments employ xVerify-0.5B-I. This is the result for the full benchmark dataset, with 95\% confidence interval}
  \label{tab:chemistry-model-table-split-2}
\end{sidewaystable*}

\section{Results \& Analysis} 

\subsection{Overall: Discriminative Power of Questions}
In undergraduate and graduate examinations, calculation-based problems possess high discriminative power, as typically only top-performing students can achieve high scores. Drawing from this analogy, we designed QCBench to serve a similar function for large language models. With a concise set of only 350 questions, it aims to distinguish genuinely capable models and establish a graded performance hierarchy. If both small and large models were to score uniformly high, the benchmark would lose its screening utility. The quantitative problems in QCBench are sourced from classic, well-established textbooks at the undergraduate and graduate levels. Solving these problems requires not only a solid understanding of fundamental chemistry concepts but also mastery of domain-specific formulas. Our test results (Table~\ref{tab:chemistry-model-table-split} and~\ref{tab:chemistry-model-table-split-2}) show that no current large language model achieves state-of-the-art performance across all subfields. For instance, some models like \textbf{GPT-oss-120B} may excel in \textit{Biochemistry}, while others like \textbf{QWQ-32B} perform better in \textit{Polymer Chemistry}. Furthermore, the results indicate that more parameters do not necessarily correlate with better performance. In fact, for large models not explicitly designed or trained for chemistry-related computations, the effects of scaling may not be apparent at all.

\paragraph{QCBench-84 vs QCBench-350}
A comparison of the results from the balanced QCBench-84 (Table \ref{tab:chemistry-model-table-split}) and the full QCBench-350 (Table \ref{tab:chemistry-model-table-split-2}) reveals a high degree of consistency in the overall performance hierarchy of the models. The relative rankings of the models remain largely stable; for instance, \textbf{o3} is the top-performing model in both datasets, and other leading models like \textbf{Grok-4} and \textbf{QwQ-32B} maintain their strong positions with only minor fluctuations in their average approximate scores (typically within 1-2 percentage points). This stability suggests that the balanced subset is a reliable proxy for general model capability.

\subsection{Challenging Sub-fields}

From the average task performance on the full benchmark dataset QCBench-350 (Table \ref{tab:chemistry-model-table-split-2}), we observe a clear hierarchy of difficulty across the chemistry sub-fields for current large language models. \textit{Analytical Chemistry} and \textit{Polymer Chemistry} present the most significant challenges, with the lowest average approximate matching scores of 25.3\% and 27.1\%, respectively. The difficulty in \textit{Analytical Chemistry} likely stems from its requirement for multi-step reasoning, precise stoichiometric calculations, and an implicit understanding of experimental procedures and data interpretation (\textit{e.g.}, titrations, spectroscopy). Similarly, \textit{Polymer Chemistry} problems often involve complex concepts such as molecular weight distributions, polymerization kinetics, and the manipulation of intricate structural formulas, which appear to be outside the core competency of most generalized models.
In stark contrast, models demonstrated the highest proficiency in \textit{Inorganic Chemistry}, achieving a notable average approximate score of 51.1\%. This suggests that the principles of inorganic chemistry, such as periodic trends, coordination compounds, and basic reaction balancing, might be better represented in the models' vast training corpora. \textit{Physical Chemistry} and \textit{Quantum Chemistry} occupy a middle ground in terms of difficulty, with average scores of 35.0\%. These fields demand a strong grasp of mathematical formalism and physical laws, indicating that while models have some capability in these areas, there is substantial room for improvement. This disparity underscores a critical insight: current LLMs are not uniformly proficient across all scientific domains, and their performance is heavily influenced by the nature and complexity of the calculations required.


\subsection{Closed-source vs. Open-source Models}
Our results reveal a nuanced landscape when comparing closed-source and open-source models. While on average, the top-tier closed-source models exhibit a performance advantage, the gap is not absolute, and leading open-source models demonstrate state-of-the-art capabilities in specific sub-fields.
Among all evaluated models, the closed-source model \textbf{o3} emerges as the overall top performer, achieving the highest average approximate score of 46.8\%. Its strength is not confined to a single area; it achieves the best performance in both \textit{General Chemistry} (50.0\%) and \textit{Inorganic Chemistry} (64.5\%), and secures the second-best position in highly challenging fields like \textit{Biochemistry} and \textit{Quantum Chemistry}. Following closely is \textbf{Grok-4}, another powerful closed-source model with an average score of 43.3\%. \textbf{Grok-4} particularly excels in the mathematically intensive field of \textit{Physical Chemistry}, where it obtains the highest score of 52.7\%, and it shares the top rank in \textit{General Chemistry}, showcasing its robust and versatile reasoning abilities.

The open-source community has produced highly competitive models that challenge the dominance of their closed-source counterparts in specialized domains. \textbf{QwQ-32B} stands out as the leading open-source model with a strong average approximate accuracy of 41.2\%. Impressively, it achieves the absolute top score in \textit{Polymer Chemistry} (37.5\%), one of the most difficult sub-fields for all models. This indicates a high degree of specialization.
Furthermore, \textbf{GPT-oss-120B} delivers an exceptional performance, nearly matching the top open-source model with an average score of 40.1\%. Its capabilities are particularly noteworthy as it achieves the highest scores across all models in both \textit{Analytical Chemistry} (40.0\%) and \textit{Biochemistry} (54.7\%), demonstrating that targeted open-source efforts can lead to state-of-the-art results. Similarly, \textbf{DeepSeek-R1} proves its mettle by obtaining the top score in \textit{Quantum Chemistry} (52.6\%). At the other end of the spectrum, models such as \textbf{Llama-3-405b-I} (18.8\%) and \textbf{Gemma-3-27B-it} (19.2\%) consistently rank among the weakest performers. Their low scores across nearly all subfields highlight the benchmark's effectiveness in differentiating model capabilities and underscore the significant challenges that quantitative chemistry poses to non-specialized large language models.

\subsection{Difficulty Levels vs. Performance}
\begin{figure}[!htb]
    \centering
    \includegraphics[width=0.9\linewidth]{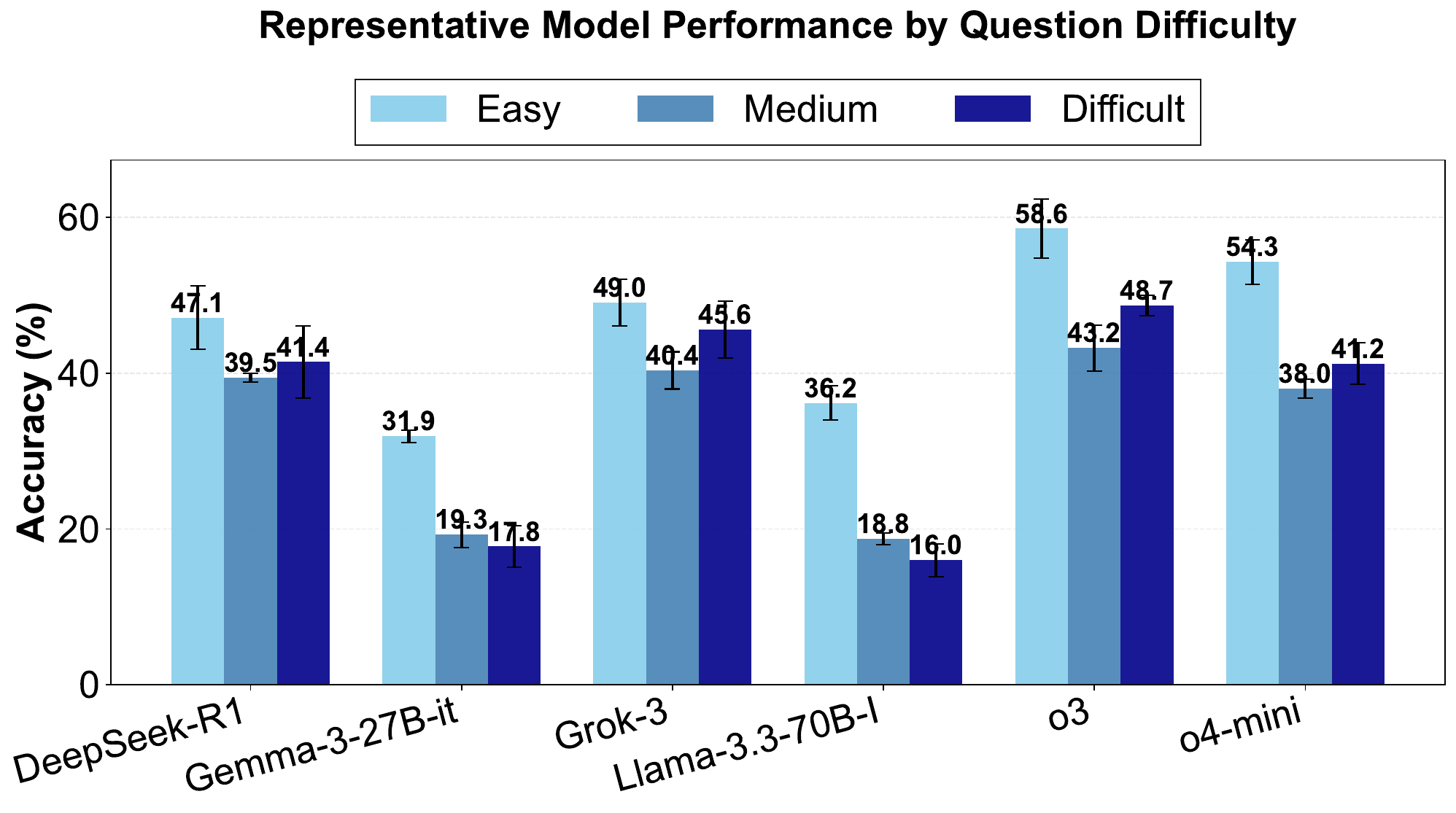}
    \caption{Accuracies on difficulty levels (Approximate Matching on QCBench-350). }
    \label{fig:difficulty_level_1}
\end{figure}

\begin{figure}[!htb]
    \centering
    \includegraphics[width=0.9\linewidth]{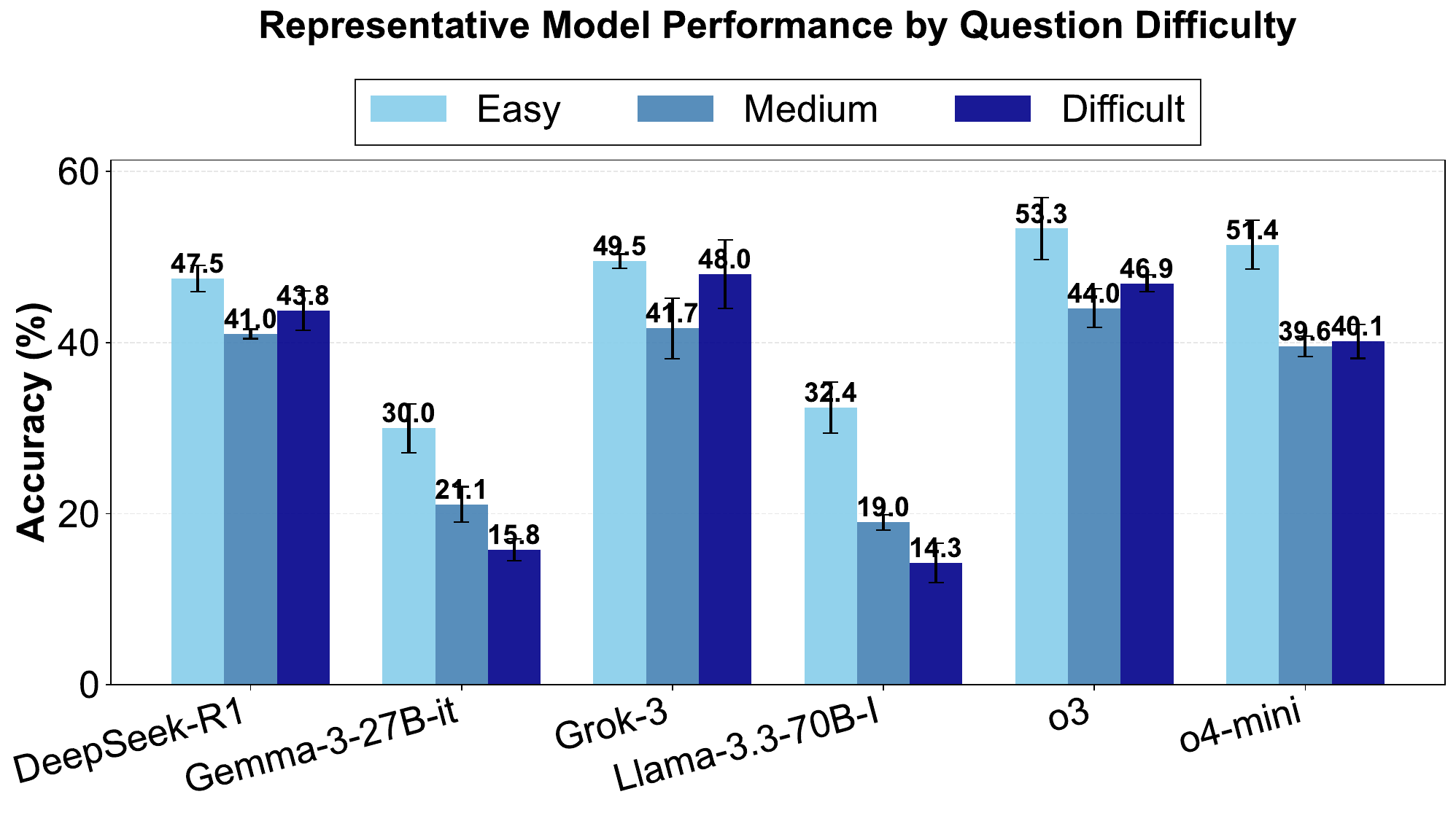}
    \caption{Accuracies on difficulty levels (xVerify on QCBench-350) }
    \label{fig:difficulty_level_2}
\end{figure}

Figures~\ref{fig:difficulty_level_1} and~\ref{fig:difficulty_level_2} illustrate the performance of representative models across three difficulty tiers (easy, medium, and difficult) by length, evaluated using Approximate Matching and the stricter xVerify, respectively. Our analysis reveals three distinct model archetypes.

\paragraph{Robust and Counter-intuitive Models} Elite models like \textbf{Grok-3} and \textbf{DeepSeek-R1} exhibit top-tier performance but display a counter-intuitive trend: their accuracy on medium and difficult questions sometimes surpasses their performance on easy ones. While this may indicate advanced reasoning capabilities, a deeper analysis suggests two contributing factors. First, our length-based difficulty heuristic may create artifacts where shorter easy questions lack sufficient context. Second, we observe a clear pattern of overthinking: these models often apply unnecessarily complex reasoning to simple problems, leading them to infer non-existent complexities and ultimately arrive at incorrect solutions.

\paragraph{Strong but Nuanced Models}
Unlike a conventional model that shows a simple decline, some models like \textbf{o3} and \textbf{o4-mini} exhibit a more complex profile. Both models achieve their highest accuracy on easy questions, as expected, with \textbf{o3} reaching an impressive 58.6\%. However, both models also perform better on difficult questions than on medium ones. For example, \textbf{o3}'s accuracy is 48.7\% on difficult tasks compared to 43.2\% on medium tasks. This indicates that while these models are well-rounded, they possess strong capabilities for handling complex problems, outperforming their results on intermediate-difficulty questions.

\paragraph{Difficulty-Sensitive Models}
Some models show a high sensitivity to question difficulty. They exhibit a clear and significant drop-off in performance as the complexity rises. For instance, \textbf{Gemma-3-27B-it}'s accuracy falls from 31.9\% on easy questions to just 17.8\% on difficult ones. Similarly, \textbf{Llama-3.3-70B-I} sees its performance halved, from 36.2\% on easy ones to 16.0\% on difficult ones. This steep, monotonic decline is characteristic of models that may lack the specialized reasoning capacity required for more complex, multi-step chemical problems.

\subsection{Recalibration of difficulty Levels by Qwen3-Max}
\begin{figure}[!htb]
    \centering
    \includegraphics[width=0.9\linewidth]{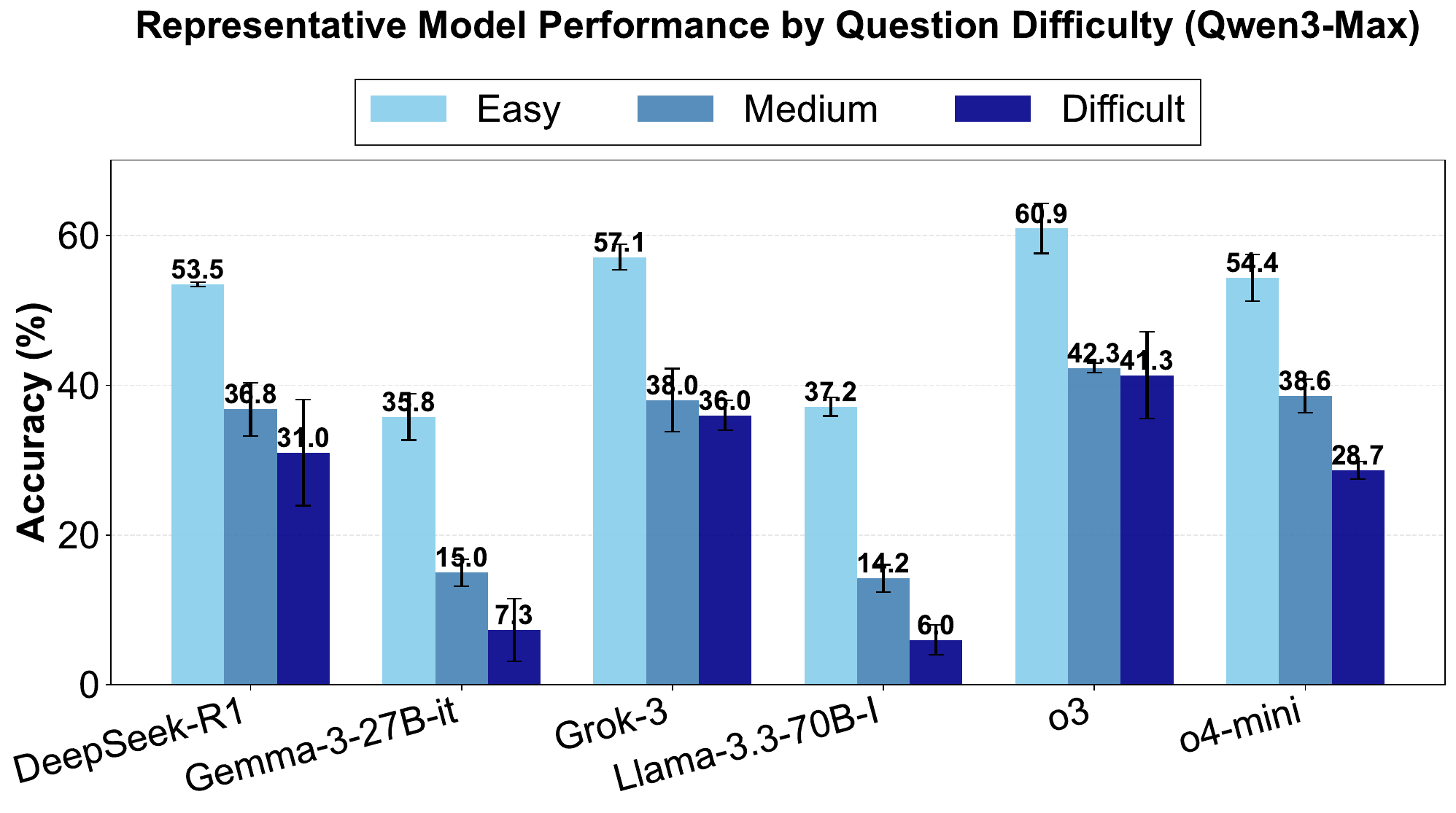}
    \caption{Accuracies on difficulty levels (Approximate Matching on QCBench-350 by Qwen3-Max). }
    \label{fig:difficulty_level_1_qwen}
\end{figure}

\begin{figure}[!htb]
    \centering
    \includegraphics[width=0.9\linewidth]{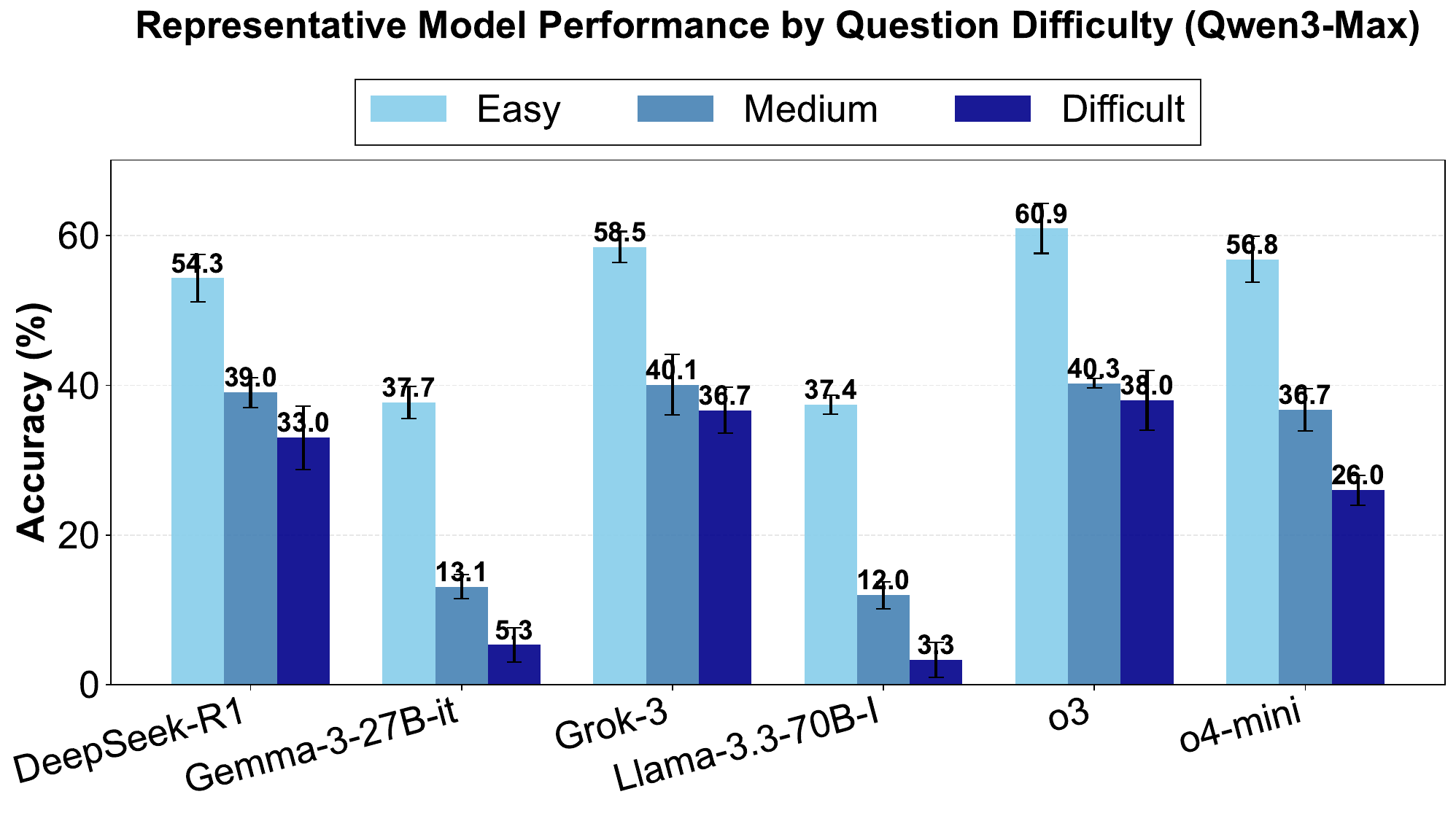}
    \caption{Accuracies on difficulty levels (xVerify on QCBench-350 by Qwen3-Max) }
    \label{fig:difficulty_level_2_qwen}
\end{figure}

Recognizing that heuristics such as problem length provide only a superficial measure of difficulty, we established a more robust and semantically meaningful categorization for our analysis. We used a state-of-the-art large language model, \textbf{Qwen3-Max}, to re-evaluate and assign a difficulty level (easy, medium, or difficult) to each question in the QCBench dataset. To ensure impartiality and prevent any potential bias from data leakage or circular evaluation, Qwen3-Max was used exclusively for this annotation task and was not included in our set of evaluated models. The performance of representative models under this new, more nuanced difficulty scheme is presented in Figures~\ref{fig:difficulty_level_1_qwen} and~\ref{fig:difficulty_level_2_qwen}.

Under the Qwen3-Max difficulty ratings, the previously observed counter-intuitive trends in elite models have largely disappeared. These models now exhibit a more predictable performance curve: they achieve their highest accuracy on easy questions and show an expected decline as the semantic difficulty increases. For example, \textbf{Grok-3} and \textbf{o3} demonstrate exceptional capability on easy tasks, with accuracies reaching 59.0\% and 57.4\%, respectively. Their performance then moderately decreases on medium (40.4\% and 42.7\%) and difficult (38.0\% for both) questions. Similarly, \textbf{DeepSeek-R1} and \textbf{o4-mini} follow this pattern, with a monotonic decrease in accuracy from easy to difficult tasks. This suggests that the previously unusual performance curves were likely an artifact of the superficial length-based difficulty metric rather than an inherent property of the models' reasoning.


\begin{table}[!htb]
\centering
\begin{adjustbox}{max width=0.92\textwidth}\footnotesize
\begin{tabular}{lcccc}
\toprule
\textbf{Class} & \textbf{Abbrev.} & \textbf{Easy} & \textbf{Medium} & \textbf{Difficult} \\
\midrule
Analytical           & AC   &    2 &     12 &    11 \\
Biochemistry         & BOC  &   11 &     10 &     3 \\
General              & GC   &    3 &     11 &     2 \\
Inorganic            & IC   &   15 &     25 &     6 \\
Physical             & PhC  &   70 &     97 &    20 \\
Polymer              & PoC  &    3 &      5 &     4 \\
Quantum              & QC   &   18 &     17 &     4 \\
\midrule
Total                &      &  122 &    178 &    50 \\
\bottomrule
\end{tabular}
\end{adjustbox}
\vspace{-0.2cm}
\caption{Distribution of question difficulties across chemistry subfields in QCBench (350 questions) by Qwen3-Max.}
\label{tab:qcbench_difficulty_qwen3}
\end{table}

\begin{table}[!htb]
\centering
\begin{adjustbox}{max width=0.92\textwidth}\footnotesize
\begin{tabular}{lcccc}
\toprule
\textbf{Class} & \textbf{Abbrev.} & \textbf{Easy} & \textbf{Medium} & \textbf{Difficult} \\
\midrule
Analytical           & AC   &    2 &      4 &     6 \\
Biochemistry         & BOC  &    7 &      5 &     0 \\
General              & GC   &    3 &      8 &     1 \\
Inorganic            & IC   &    3 &      9 &     0 \\
Physical             & PhC  &    6 &      4 &     2 \\
Polymer              & PoC  &    3 &      5 &     4 \\
Quantum              & QC   &    7 &      3 &     2 \\
\midrule
Total                &      &   31 &     38 &    15 \\
\bottomrule
\end{tabular}
\end{adjustbox}
\vspace{-0.2cm}
\caption{Distribution of question difficulties across chemistry subfields in QCBench-balanced (84 questions) by Qwen3-Max.}
\label{tab:qcbench_balanced_difficulty_qwen3}
\end{table}

\begin{figure}[!htb]
    \centering
    \includegraphics[width=0.9\linewidth]{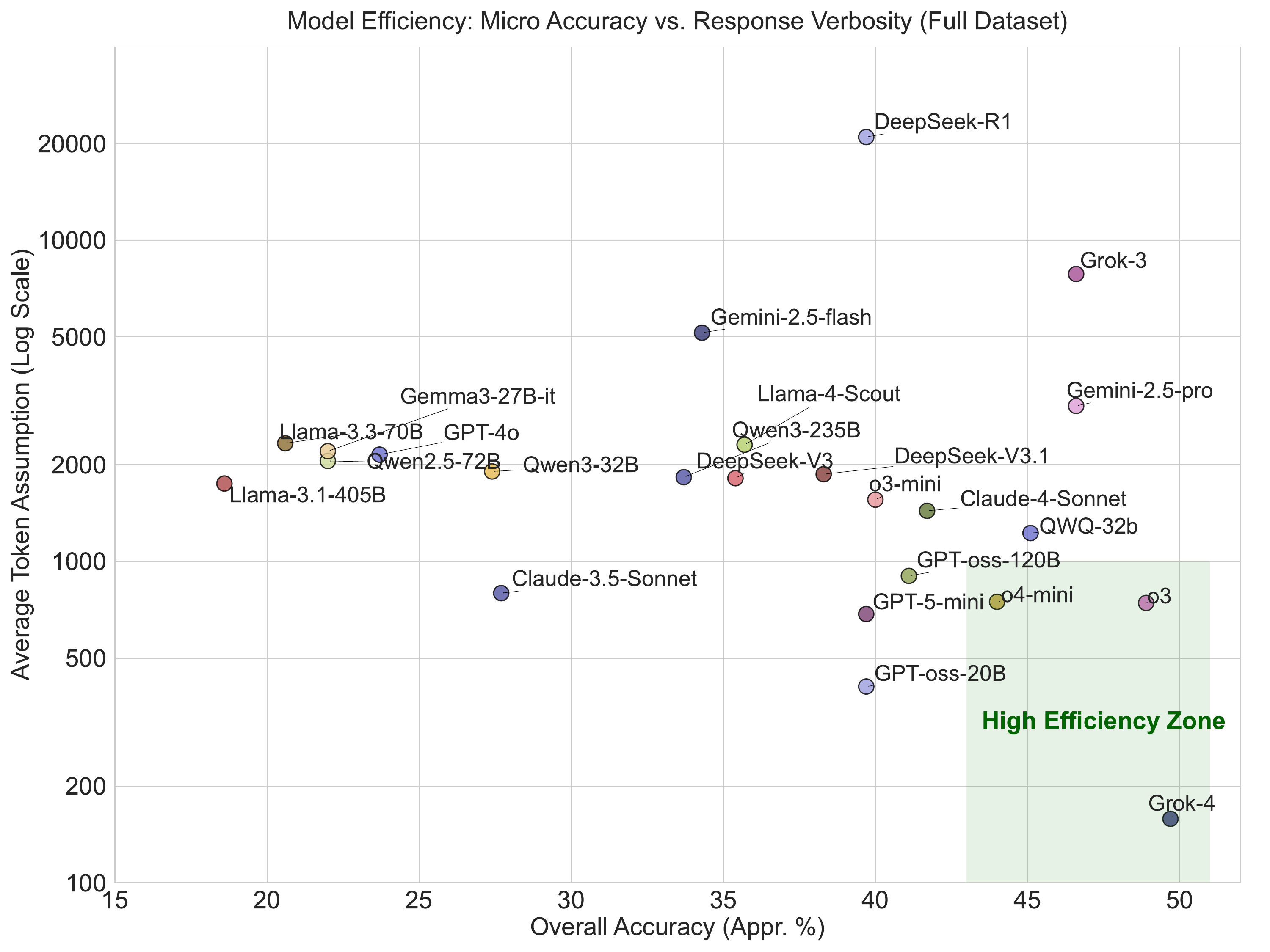}
    \caption{Model Accuracy vs. Response Verbosity: A Tale of Efficiency. }
    \label{fig:performance_vs_eff}
\end{figure}

\subsection{Model Accuracy, Response Verbosity and Runtime Analysis}
The relationship between model accuracy, response verbosity, and runtime reveals a multi-faceted interplay where efficiency is not a single metric but a trade-off between computational cost and token output. We compute the micro accuracy, which is the total accuracy instead of the macro accuracy (class-wise average accuracy). All statistics are presented in the Table~\ref{tab:token_usage_runtime}. As illustrated in Figure~\ref{fig:performance_vs_eff}, models adopt distinct strategies, leading to varied performance profiles.
A clear quadrant of high token-efficiency is dominated by models that achieve top-tier accuracy with remarkable brevity. \textbf{Grok-4} is the quintessential example, delivering the highest accuracy (49.7\%) with unparalleled conciseness (158 average tokens). This suggests a highly optimized and confident reasoning process. However, this token-miserliness comes at a steep computational cost, as it has one of the longest runtimes (154s), which might account for its internal reasoning mode. Similarly, \textbf{o3} pairs the second-highest accuracy (48.9\%) with very low verbosity (743 tokens), defining a group of models that prioritize precise and direct answers, even if it requires more processing time.
In stark contrast, another group of high-performing models adopts a verbose but powerful strategy. \textbf{Gemini-2.5-pro} and \textbf{Grok-3}, for instance, achieve high accuracy (both 46.6\%) but generate substantially longer responses (3,050 and 7,854 tokens, respectively). This verbosity is not necessarily a flaw but may reflect a more transparent \textbf{chain-of-thought} or explanatory reasoning style, providing detailed steps at the cost of higher token usage.
Finally, the chart highlights models where verbosity becomes a hallmark of inefficiency. Models like \textbf{Llama-3.3-70B} and \textbf{Gemma-3-27B-it} occupy the upper-left region, pairing low accuracy with high token counts. \textbf{DeepSeek-R1} represents an extreme outlier; while its accuracy is respectable (39.7\%), its average token consumption is an order of magnitude higher than any other model, making it impractical for most applications despite its capabilities. This analysis demonstrates that there is no single ideal profile; the optimal model choice depends on whether the user prioritizes token economy (\textbf{Grok-4}, \textbf{o3}), fast runtimes (\textbf{GPT-oss-20B}), or a detailed, albeit lengthy, reasoning process (\textbf{Gemini-2.5-pro}).

\section{Ablation Studies}

\subsection{Split and Sum}
There are only two sum-type questions: one from QCBench (index 196) and another from QCBench (index 338). For the split-type questions, they correspond to indices 12 and 13 in SciBench, which were not curated by us. First, since we only evaluate models without any additional training on these items, the risk of data leakage does not exist. Second, we observe that the performance on the original problems and their corresponding split or sum versions remains identical, indicating that such augmentations do not introduce artifacts or alter the difficulty of the tasks.

\subsection{Tolerance and Stability Study}
\begin{figure}[!htb]
    \centering
    \includegraphics[width=0.8\linewidth]{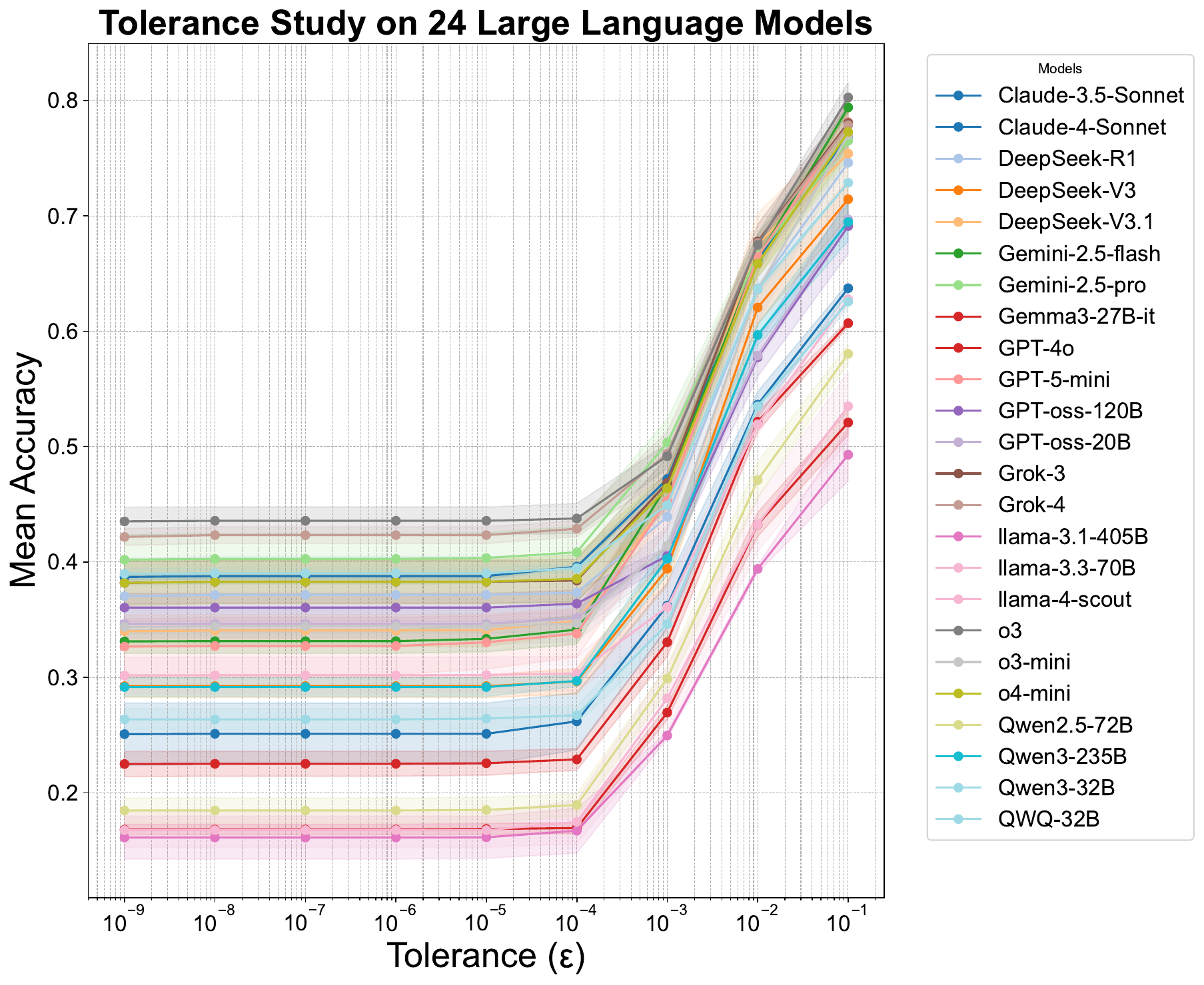}
    \caption{Tolerance Level vs. Accuracy}
    \label{fig:tolerance_vs_accuracy}
\end{figure}
To provide a granular analysis of the models' raw numerical precision and performance stability, we conducted a comprehensive tolerance ablation study across all 24 models. This analysis evaluates how the \textbf{mean accuracy}, calculated over multiple independent runs, varies as the acceptable margin of relative error ($\epsilon$) is adjusted. Unlike the primary results presented in Table~\ref{tab:chemistry-model-table-split} and~\ref{tab:chemistry-model-table-split-2}, which may involve answer normalization, this study strictly evaluates the models' unprocessed numerical outputs. This methodology allows us to distinguish between minor precision errors and more significant reasoning failures, providing a deeper understanding of model behavior.

As illustrated in Figure~\ref{fig:tolerance_vs_accuracy}, the results reveal a clear performance stratification and a consistent trend across all models. At stringent tolerances, model performance is stable but clearly layered. A distinct top tier emerges, with models like \textbf{o3} and \textbf{Grok-4} achieving mean accuracies above \textbf{42\%}. In contrast, a significant portion of the models, including \textbf{Llama-3.1-405B} and \textbf{Gemma3-27B-it}, score below \textbf{17\%}, establishing a wide initial performance gap. The consistently low standard deviation across all models in this range (typically $\epsilon< 0.02$) indicates that their precision, whether high or low, is highly consistent across runs.

\begin{figure}[!htb]
    \centering
    \includegraphics[width=0.8\linewidth]{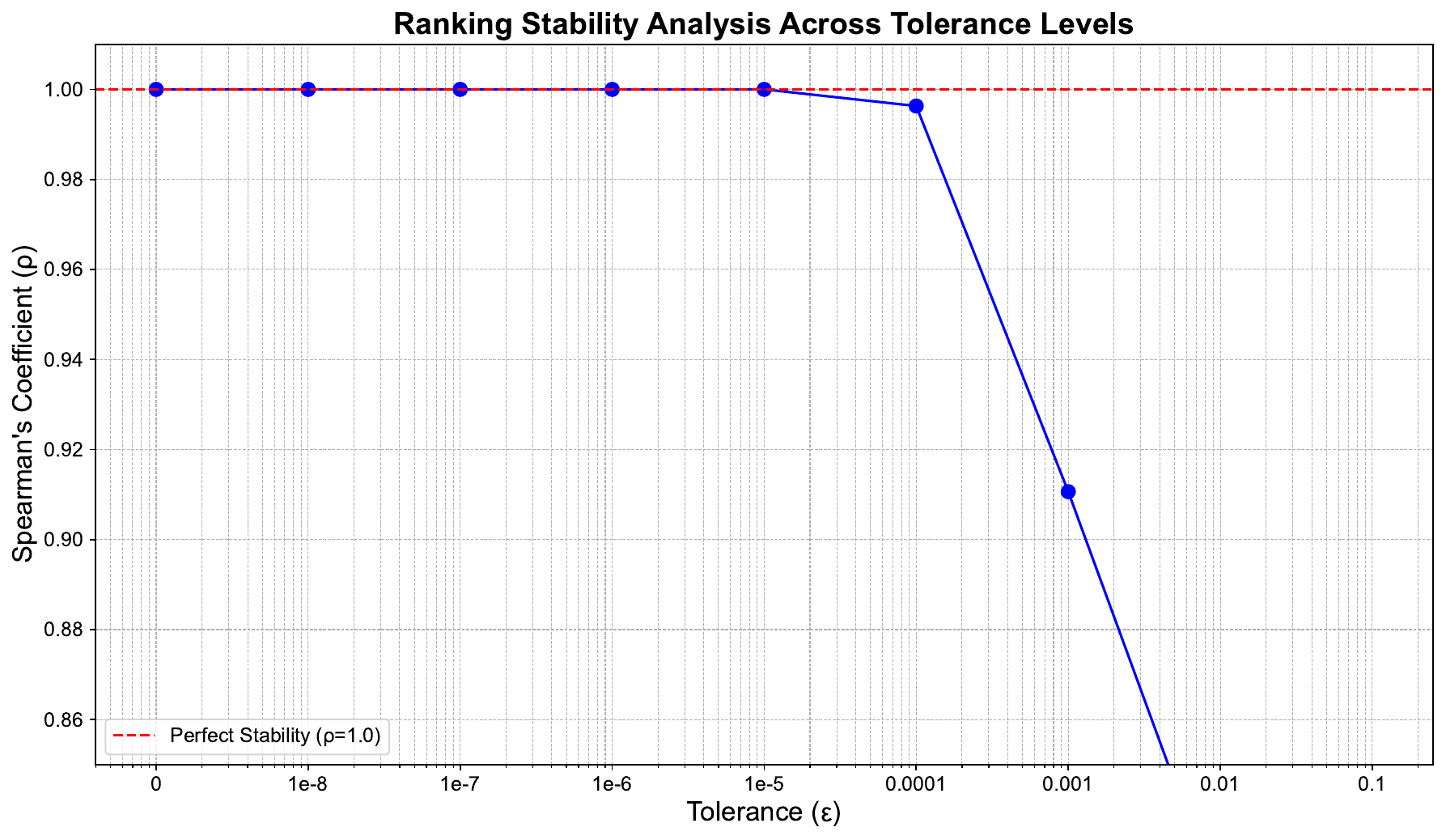}
    \caption{Model Ranking Stability}
    \label{fig:rank_stability}
\end{figure}

A dramatic and universal increase in accuracy occurs as the tolerance is relaxed. The magnitude of this improvement is substantial. For instance, \textbf{Grok-3} nearly doubles from $\approx$38\% to $\approx$68\%. The effect is even more pronounced for other models; the accuracy of \textbf{Gemini-2.5-flash}, for example, increases by over 100\% (from $\approx$33\% to $\approx$66\%), while a lower-tier model like \textbf{Llama-3.3-70B} sees its performance nearly triple (from $\approx$17\% to $\approx$43\%). At the most lenient tolerance, the performance hierarchy largely holds, but the gap widens. Top models like \textbf{o3} now exceed \textbf{80\%} mean accuracy, while the lowest-performing models still struggle to reach \textbf{50\%}. This demonstrates that while leniency helps all models, it does not erase the fundamental differences in their quantitative reasoning capabilities. We also observe that the ranking stability is relatively high according to Spearman's Coefficient (Figure~\ref{fig:rank_stability}).

\begin{table}[h]
\centering
\caption{Token usage and runtime.}
\label{tab:token_usage_runtime}
\begin{tabular}{lcccc}
\toprule[1pt]
Model Name & Runtime (s) & \#Tokens &Micro Acc & \#Samples \\
\midrule 
Gemini-2.5-flash & 8.68 & 5,155 & 0.343 & 350 \\
Claude-3.5-Sonnet & 10.27 & 797 & 0.277 & 350 \\
GPT-4o & 10.68 & 2,151 & 0.237 & 350 \\
GPT-oss-20B & 10.94 & 408 & 0.397 & 350 \\
Claude-4-Sonnet & 10.97 & 1,438 & 0.417 & 350 \\
GPT-oss-120B & 12.75 & 902 & 0.411 & 350 \\
Llama-4-Scout & 13.77 & 2,309 & 0.357 & 350 \\
Qwen2.5-72B & 16.17 & 2,056 & 0.220 & 350 \\
Llama-3.3-70B & 19.43 & 2,333 & 0.206 & 350 \\
o4-mini & 23.16 & 749 & 0.440 & 350 \\
Qwen3-32B & 24.16 & 1,907 & 0.274 & 350 \\
Gemma3-27B-it & 24.82 & 2,205 & 0.220 & 350 \\
DeepSeek-V3.1 & 25.42 & 1,870 & 0.383 & 350 \\
Llama-3.1-405B & 28.84 & 1,749 & 0.186 & 350 \\
DeepSeek-V3 & 28.90 & 1,817 & 0.354 & 350 \\
o3-mini & 31.79 & 1,558 & 0.400 & 350 \\
GPT-5-mini & 39.59 & 686 & 0.397 & 350 \\
Grok-3 & 41.15 & 7,854 & 0.466 & 350 \\
o3 & 42.58 & 743 & 0.489 & 350 \\
Gemini-2.5-pro & 64.74 & 3,050 & 0.466 & 350 \\
Grok-4 & 154.47 & 158 & 0.497 & 350 \\
Qwen3-235B & 210.20 & 1,830 & 0.337 & 350 \\
QWQ-32B & 261.71 & 1,226 & 0.451 & 350 \\
DeepSeek-R1 & 862.85 & 20,970 & 0.397 & 350 \\
\bottomrule[1pt]
\end{tabular}
\end{table}

\subsection{Approximate Verification vs. Exact Matching}

An analysis of the benchmark's approximate accuracy (Appr.) and the exact matching enabled by xVerify reveals a crucial dynamic: for many advanced models, the verification process functions as a powerful self-correction and reasoning-enhancement tool. This is a significant shift from the notion of a verifier merely acting as a restrictive filter. The data in Table \ref{tab:chemistry-model-table-split-2} shows that numerous top-performing models achieve a higher score under the stricter xVerify metric than with the more lenient approximate matching.
This trend is exemplified by several leading models. For instance, the closed-source model \textbf{Grok-4} sees its average performance increase from 43.3\% (Appr.) to 46.1\% (xVerify). Even more striking are the gains observed in the open-source domain: \textbf{DeepSeek-R1}'s accuracy jumps from 38.4\% to 42.7\%, and \textbf{GPT-oss-120B} improves from 40.1\% to 41.9\%. This performance uplift indicates that these models possess a sophisticated internal reasoning capability. When prompted to verify their work, they can re-trace their steps, identify initial fallacies—be they calculation errors or flawed logic—and converge on the correct, verifiable answer. The verification step is not just about formatting; it is an active problem-solving phase. For example, at the index 166 for computing the free energy difference giving the dissociation constant for the particular protein dimer, \textbf{Grok-4} gives the answer -34.2, where the groundtruth is 34.2. Although a direct numerical comparison would flag the answer as incorrect, xVerify’s more sophisticated evaluation correctly identifies it as valid, demonstrating its ability to handle chemically meaningful equivalences.
However, this self-correction capability is not universal and serves as a key differentiator. Less capable models, such as \textbf{Llama-3-405b-I} (18.8\% Appr. vs. 17.8\% xVerify) and \textbf{Gemma-3-27B-it} (19.2\% vs. 18.0\%), tend to see their scores decrease after verification. This suggests that their initial correct answers might be a result of lucky guesses or superficial pattern matching, and their underlying reasoning is not robust enough to withstand the scrutiny of a verification process. One typical example is the index 140 of the QCBench-350, where the answer is $3.548*10^{-27}$. The predicted answer by \textbf{Gemma3-27B-it} is $3.54*10^{-24}$, If we directly compare the difference, it's small because of the scale. Therefore, the gap between Appr. and xVerify is no longer an indicator of a verifier's limitation. Instead, a positive gap (where xVerify > Appr.) serves as a strong signal of a model's advanced, reliable reasoning and self-correction abilities, separating models that can genuinely solve a problem from those that can only approximate a plausible-sounding answer.

\subsection{Different Temperatures}
We observe that performance does not vary substantially across temperature choices. 
At temperature $T=0$, decoding becomes deterministic (greedy), yielding approximate unique outputs but not higher accuracy. 
Non-zero temperatures ($T>0$) introduce randomness, which can slightly affect Approximate Match and xVerify scores but without clear monotonic trends. That is the main reason we chose the temperature $T= 0.1$ in our experiment settings.
\begin{table}[!htb]
\centering \scriptsize
\caption{Performance (Micro Accuracy) under different temperature settings with fixed $top\_p=1.0$.}
\begin{tabular}{lcccccc}
\toprule
\multirow{2}{*}{Model} & \multicolumn{2}{c}{T = 1} & \multicolumn{2}{c}{T = 0.5} & \multicolumn{2}{c}{T = 0} \\
\cmidrule(lr){2-3} \cmidrule(lr){4-5} \cmidrule(lr){6-7}
 & Approx.~Match & xVerify & Approx.~Match & xVerify & Approx.~Match & xVerify \\
\midrule
o4-mini & $0.517 \pm 0.012$ & $0.460 \pm 0.004$ & $0.520 \pm 0.008$ & $0.446 \pm 0.000$ & $0.516 \pm 0.006$ & $0.453 \pm 0.010$ \\
GPT-oss-120B & $0.420 \pm 0.004$ & $0.449 \pm 0.008$ & $0.416 \pm 0.010$ & $0.449 \pm 0.012$ & $0.413 \pm 0.002$ & $0.446 \pm 0.020$ \\
\bottomrule
\end{tabular}
\end{table}

\subsection{Thinking vs. No Thinking}

\begin{figure}[!htb]
    \centering
    \includegraphics[width=\linewidth]{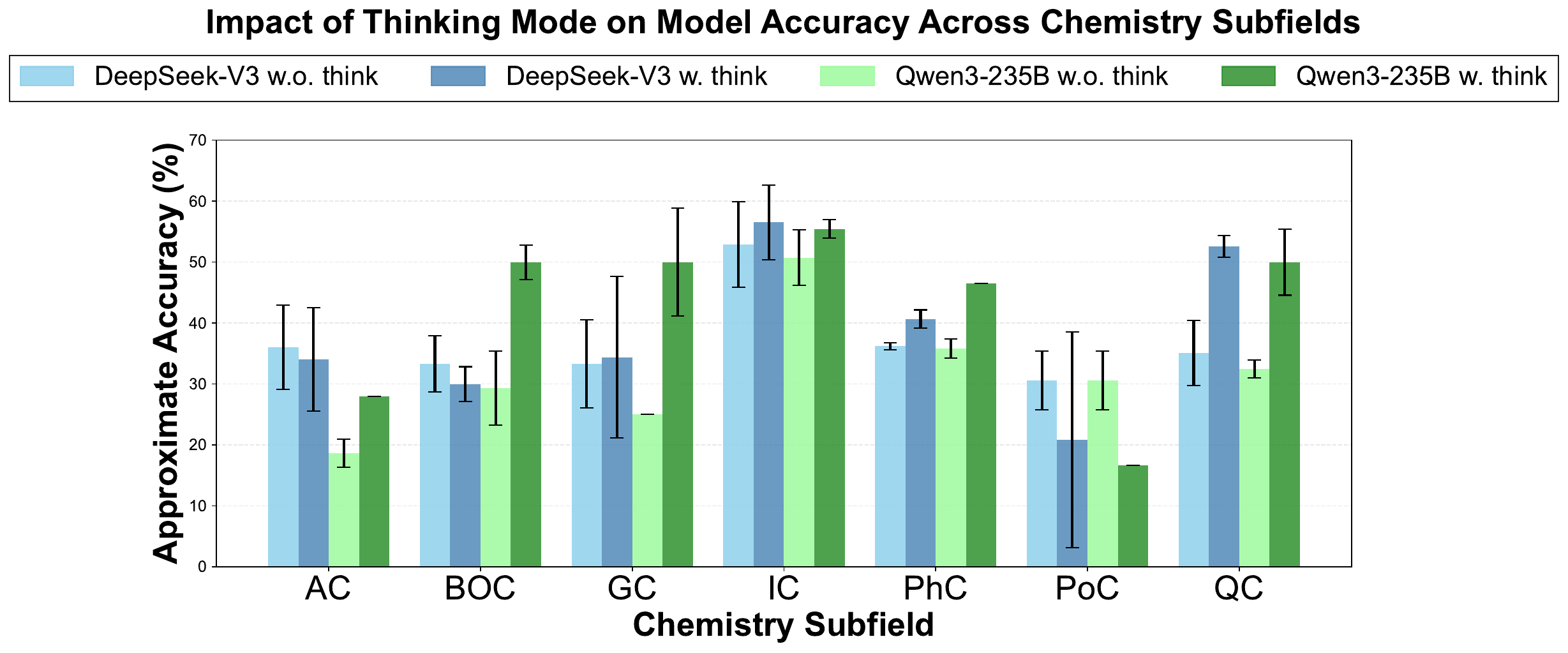}
    \vspace{-0.2cm}
    \caption{Thinking versus No Thinking (Approx). }
    \label{fig:think_vs_no_think}
\end{figure}

\begin{figure}[!htb]
    \centering
    \includegraphics[width=\linewidth]{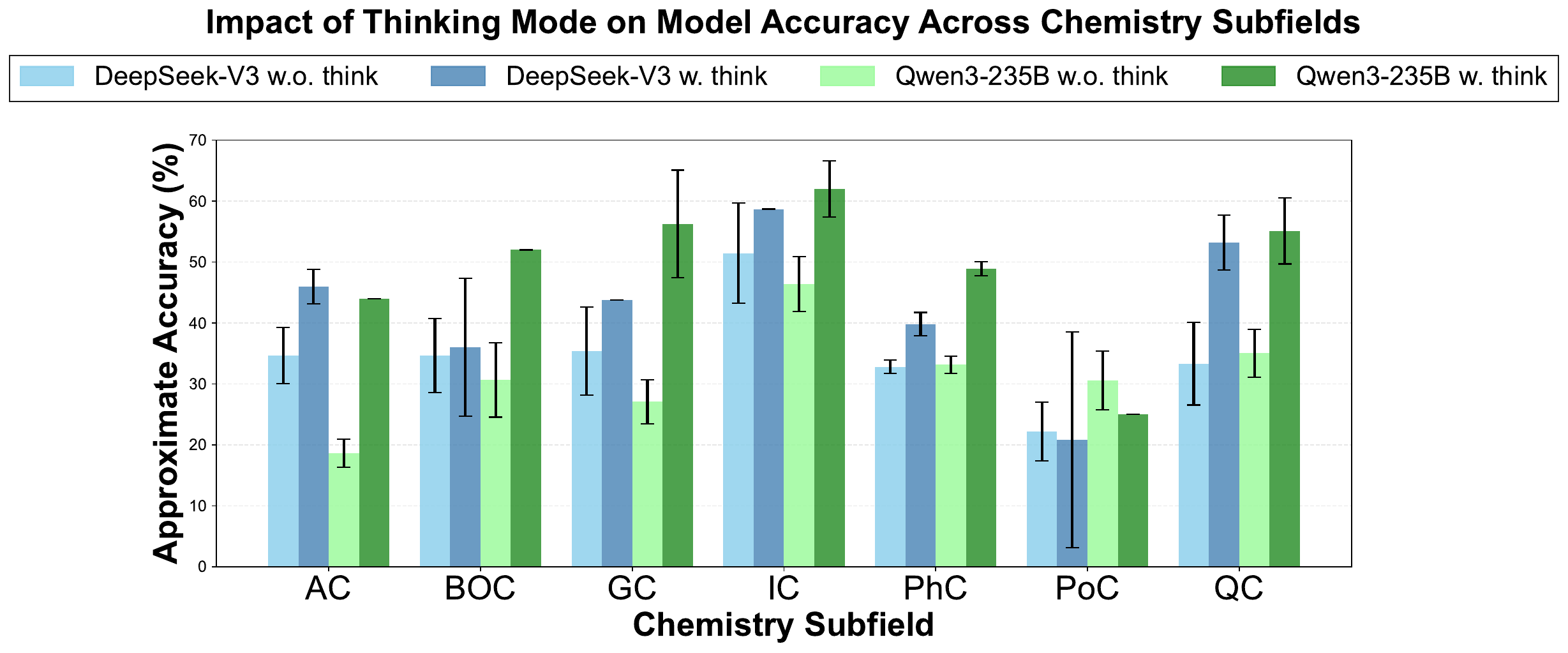}
    \vspace{-0.2cm}
    \caption{Thinking versus No Thinking (xVerify). }
    \label{fig:think_vs_no_think_xverify}
\end{figure}

To investigate the impact of explicit chain-of-thought prompting on model performance, we conducted a comparative analysis of two representative models, \textbf{DeepSeek-V3} and the open-source \textbf{Qwen3-235B}, with and without thinking mode enabled. This mode explicitly instructs the model to externalize its step-by-step reasoning process before providing a final answer. The results, presented in Figure~\ref{fig:think_vs_no_think} and~\ref{fig:think_vs_no_think_xverify}, reveal that the benefits of this mode are highly model-dependent and vary significantly across different chemistry subfields.

For \textbf{DeepSeek-V3}, enabling the thinking mode provides a substantial and almost universally positive performance uplift. Across nearly all seven subfields, with thinking configuration (typically DeepSeek-R1) consistently outperforms the DeepSeek-V3 baseline. The most dramatic improvements are observed in domains requiring complex calculations, such as \textit{Inorganic Chemistry} (IC), where accuracy increases from approximately 52\% to 60\%, and \textit{Quantum Chemistry} (QC), which sees a gain from 34\% to over 50\%. This strongly suggests that for a capable, closed-source model like DeepSeek-V3, externalizing the reasoning process is a critical mechanism for ensuring accuracy in intricate logical and mathematical deductions. Similar patterns are applied to the \textbf{Qwen3-235B} model.


\paragraph{The Nature of Thinking vs. Mere Verbosity:}
These results confirm that the value of the thinking mode is fundamentally different from simply generating a longer, more verbose answer. At its core, this mode provides a structured and traceable reasoning process, allowing the model to deconstruct a complex problem into manageable sub-steps. The dramatic improvement of Qwen3-235B in \textit{Biochemistry}, for example, is not a result of increased text output but of its chain-of-thought process successfully executing the required multi-step logic. This is qualitatively different from the \textit{pseudo-verbosity} of a non-thinking model, which might generate excessive, unstructured text for a simple query without any improvement in accuracy. The cases where thinking mode fails (\textit{e.g.}, Qwen3-235B in \textit{Polymer Chemistry}) further highlight this distinction, suggesting that a flawed reasoning chain, no matter how detailed, cannot compensate for underlying knowledge gaps.

\section{Ranking Decoupling Across Chemical Benchmarks}

To contextualize our findings and validate the unique contribution of QCBench, we compared the model performance hierarchy with that of ChemBench~\cite{mirza2024large}, a well-established, broader benchmark for chemical capabilities. Instead of finding consistent rankings, our analysis reveals a significant and insightful \textbf{ranking decoupling} between the two benchmarks, highlighting the specialized nature of the skills measured by QCBench. The comparative analysis for models present in both our full QCBench-350 dataset (Table~\ref{tab:chemistry-model-table-split-2}) and the ChemBench leaderboard is summarized in Table~\ref{tab:rank_decoupling}.

\begin{table}[h!]
\centering
\caption{Demonstration of ranking decoupling between QCBench (quantitative focus) and ChemBench (broad focus). The stark drop in ranks highlights the difference between general chemical knowledge and quantitative reasoning ability. }
\label{tab:rank_decoupling}
\begin{tabular}{l|cc|cc}
\toprule
\multirow{2}{*}{\textbf{Model}} & \multicolumn{2}{c|}{\textbf{QCBench}} & \multicolumn{2}{c}{\textbf{ChemBench~\cite{mirza2024large}}} \\
& \textbf{Rank} & \textbf{Appr. Acc.} & \textbf{Rank} & \textbf{Overall Acc.} \\
\midrule
o1 / o3 & \textbf{1st} & 0.47 & \textbf{1st} & 0.64 \\
Claude-3.5 Sonnet & 19th & 0.27 & \textbf{3rd} & 0.62 \\
GPT-4o & 16th & 0.35 & 4th & 0.61 \\
Llama-3.3-70B-Instruct & 22nd & 0.19 & 7th & 0.52 \\
\bottomrule
\end{tabular}
\end{table}

The results are striking. With the exception of the top-performing \textbf{o3} model, there is a dramatic divergence in rankings. Most notably, \textbf{Claude-3.5 Sonnet}, a top-3 model on ChemBench, plummets to 19th place on QCBench, with its absolute accuracy dropping from 62\% to a mere 27.4\%. Similarly, other highly-ranked models on ChemBench, such as \textbf{GPT-4o} and \textbf{Llama-3.3-70B-I}, experience a severe drop in both rank and performance on our benchmark. This decoupling phenomenon strongly supports our central thesis. Performance on a broad-domain benchmark like ChemBench reflects a model's ability as a comprehensive chemical knowledge base, excelling at qualitative, factual, and simple predictive tasks. In contrast, QCBench is specifically designed to evaluate a model's capability as a \textit{Chemical Computational Engine,} a skill that demands rigorous, multi-step mathematical reasoning. The stark performance drop demonstrates that these two capabilities are not interchangeable. A model can possess vast chemical knowledge yet falter when required to perform complex calculations. Therefore, this analysis validates QCBench not as a mere alternative to existing benchmarks, but as an essential and orthogonal diagnostic tool. It successfully isolates and quantifies a core, yet often obscured, dimension of scientific intelligence, revealing critical weaknesses in quantitative reasoning that are masked by strong performance in other areas.

\section{Conclusion}

In this work, we introduced QCBench, a benchmark meticulously designed to move beyond qualitative assessments and rigorously probe the quantitative reasoning of LLMs, which is a critical, yet underexplored facet of chemical intelligence. Our comprehensive evaluation using this framework has yielded several critical insights that challenge common assumptions in the field.
We confirmed a profound gap between the linguistic fluency of LLMs and their computational accuracy, with domains like \textit{Analytical} and \textit{Polymer Chemistry} acting as key frontiers. Furthermore, our findings demonstrate that top-tier performance is not merely a function of model scale, but a nuanced interplay between specialized reasoning strengths and computational efficiency. Perhaps most significantly, our analysis of the verification gap reveals a crucial methodological insight: for elite models, strict verification can act as a reasoning enhancer rather than a penalty, suggesting that evaluation tools themselves must co-evolve with the models they assess.
Ultimately, QCBench offers more than a static leaderboard. It provides a dynamic diagnostic framework that pinpoints specific weaknesses and guides the targeted development of more scientifically robust and accurate AI. This work lays the groundwork for future research into domain-specific fine-tuning, advanced self-correction mechanisms, and a deeper understanding of emergent computational capabilities in scientific AI.

\section{Author Contributions}
Jiaqing Xie is responsible for data curation, drafting the manuscript, visualization, and computational experiments. Weida Wang is responsible for coding, drafting the manuscript, computational experiments, and visualization. Ben Gao is responsible for drafting the manuscript, and data curation. Zhuo Yang is responsible for drafting the manuscript. Shufei Zhang is responsible for technical guidance. Tianfan Fu is responsible for paper proofreading. Yuqiang Li is responsible for supporting the whole project.

\section{Conflict of Interest Statement}
The authors declare no competing financial interests.

\section{Acknowledgements} 
Tianfan Fu is supported by Nanjing University International Collaboration Initiative and Distinguished Overseas Young Talents.

\section{Data and Software Availability Statement}
All data is available at \href{https://huggingface.co/datasets/jiaxie/QCBench}{https://huggingface.co/datasets/jiaxie/QCBench}. Code is available at \href{https://github.com/jiaqingxie/QCBench}{https://github.com/jiaqingxie/QCBench}. The API key should have access to the models.

\bibliography{achemso-demo}

\section*{Table of Contents (TOC) Graphic}

\begin{center}
\includegraphics[width=0.92\linewidth]{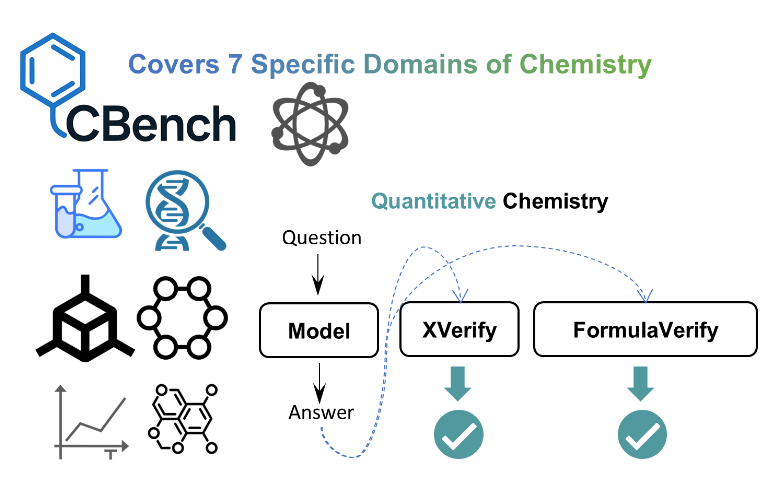}
\end{center}

\end{document}